\documentclass[lettersize,journal]{IEEEtran}
\usepackage{amsmath,amsfonts}
\usepackage{algorithmic}
\usepackage{algorithm}
\usepackage{array}
\usepackage[caption=false,font=normalsize,labelfont=sf,textfont=sf]{subfig} 

\usepackage{textcomp}
\usepackage{stfloats}
\usepackage{url}
\usepackage{verbatim}
\usepackage{graphicx}
\usepackage{cite}
\usepackage[hidelinks]{hyperref}
\usepackage{amssymb}
\usepackage{booktabs}
\usepackage{xcolor}
\usepackage{multirow} 
\usepackage{diagbox}
\usepackage{anyfontsize}
\usepackage{float}
\usepackage{colortbl}
\usepackage{enumitem}
\setlist{nosep}
\usepackage{orcidlink}

\newcommand{\blue}[1]{#1}
\newcommand{\bluex}[1]{#1}

\definecolor{lightgray}{gray}{0.9}

\usepackage{pifont}
\newcommand{\cmark}{\ding{51}}  
\newcommand{\xmark}{\ding{55}}

\begin{document}

\title{CPAM: Context-Preserving Adaptive Manipulation for Zero-Shot Real Image Editing}

\author{Dinh-Khoi Vo\orcidlink{0000-0001-8831-8846},
        Thanh-Toan Do\orcidlink{0000-0002-6249-0848},
        Tam V. Nguyen\orcidlink{0000-0003-0236-7992},
        Minh-Triet Tran\orcidlink{0000-0003-3046-3041},
        Trung-Nghia Le\orcidlink{0000-0002-7363-2610}
\thanks{Dinh-Khoi Vo, Minh-Triet Tran, and Trung-Nghia Le are with the Faculty of Information Technology, University of Science, Ho Chi Minh City, Vietnam and with Vietnam National University, Ho Chi Minh City, Vietnam (e-mail: vdkhoi@selab.hcmus.edu.vn, \{tmtriet, ltnghia\}@fit.hcmus.edu.vn).}
\thanks{Thanh-Toan Do is with the Faculty of Information Technology, Monash University, Melbourne, Victoria, Australia (e-mail: toan.do@monash.edu).}
\thanks{Tam V. Nguyen is with the Department of Computer Science, University of Dayton, Dayton, Ohio, US (e-mail: tamnguyen@udayton.edu).}
\thanks{Corresponding author: Trung-Nghia Le.}
}

\markboth{IEEE Transactions on Multimedia,~Vol.~XX, No.~X, Month~20XX}%
{Vo \MakeLowercase{\textit{et al.}}: CPAM: Context-Preserving Adaptive Manipulation for Zero-Shot Real Image Editing}



\maketitle

\begin{abstract}
Editing natural images using textual descriptions in text-to-image diffusion models remains a significant challenge, particularly in achieving consistent generation and handling complex, non-rigid objects. Existing methods often struggle to preserve textures and identity, require extensive fine-tuning, and exhibit limitations in editing specific spatial regions or objects while retaining background details. This paper proposes Context-Preserving Adaptive Manipulation (CPAM), a novel zero-shot framework for complicated, non-rigid real image editing. Specifically, we propose a preservation adaptation module that adjusts self-attention mechanisms to preserve and independently control the object and background effectively. This ensures that the objects' shapes, textures, and identities are maintained while keeping the background undistorted during the editing process using the mask guidance technique. Additionally, we develop a localized extraction module to mitigate the interference with the non-desired modified regions during conditioning in cross-attention mechanisms. We also introduce various mask-guidance strategies to facilitate diverse image manipulation tasks in a simple manner. \blue{CPAM can be seamlessly integrated with multiple diffusion backbones, including SD1.5, SD2.1, and SDXL, demonstrating strong generalization across different model architectures.} Extensive experiments on our newly constructed Image Manipulation BenchmArk (IMBA), a robust benchmark dataset specifically designed for real image editing, demonstrate that our proposed method is the preferred choice among human raters, outperforming existing state-of-the-art editing techniques. \blue{The source code and data will be publicly released at the project page: \url{https://vdkhoi20.github.io/CPAM}.}

\end{abstract}


\section{Introduction}

\label{sec:intro}

\begin{figure*}[t!]
  \includegraphics[width=\textwidth]{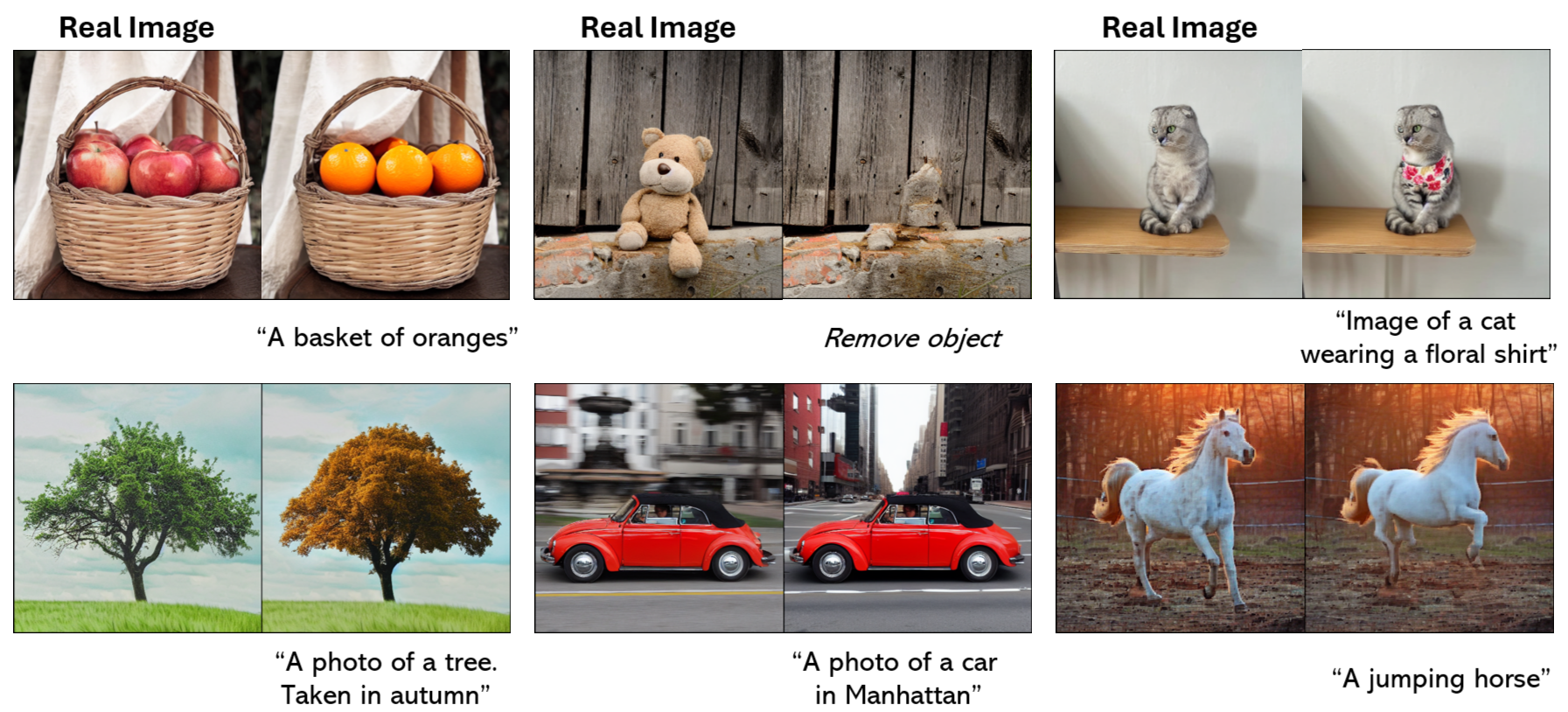}
  \vspace{-5mm}
  \caption{For each image pair, given a real image (left) and a text prompt, our method (right) facilitates \textit{zero-shot text-guided editing} without requiring fine-tuning of Stable Diffusion. Our results exhibit complex, non-rigid, consistent, and faithful editing while preserving the structure and scene layout in the original image. Our proposed method addresses various image editing tasks, including object replacement (left column), object removal and background alteration (middle column), the addition of new consistent items, and changes in object pose/view (right column).}
  \label{fig:overall}
  \vspace{-3mm}
\end{figure*}

\IEEEPARstart{R}{ecent} advancements in text-to-image (T2I) generation ~\cite{ramesh2021zeroshot,NEURIPS2021_49ad23d1,nichol2022glide,yu2022scaling,ramesh2022hierarchical,NEURIPS2022_ec795aea,blackforest2024flux,esser2403scaling} have marked significant milestones, especially with large-scale diffusion models~\cite{rombach2022highresolution} that excel in creating diverse and high-quality images from text prompts. These models have opened new avenues for text-conditioned image editing~\cite{hertz2022prompt,Tumanyan_2023_CVPR,parmar2023zeroshot}. 
 
Real image editing typically aims to produce multiple images of different complex, non-rigid objects or characters that resemble the original targeted object while also ensuring a perfect reconstruction of the background~\cite{khoi_chi2024,wallace2023edict,pan2023effective,parmar2023zeroshot}. However, this presents notable challenges. Text-guided editing of a real image using state-of-the-art diffusion models~\cite{Kim_2022_CVPR} requires inverting the given image, which involves finding an initial latent noise that accurately reconstructs the input image~\cite{deutch2024turboedit,garibi2024renoise,han2024proxedit,huberman2024edit,ju2023direct} while preserving the model’s editing capabilities. Editing an image from that latent noise often results in losing original textures and identity, leading to a different image. Additionally, existing methods are limited in editing specific objects within images, as they often focus on the most salient objects. This limitation arises from training diffusion models~\cite{rombach2022highresolution} on image-captioning datasets~\cite{schuhmann2021laion,schuhmann2022laion}, which may lack detailed descriptions of text prompts for real-world images. Thus, pre-trained Stable Diffusion (SD) is unable to focus on specific regions and instead operates on the overall image. Furthermore, real-world images often contain multiple objects and complex interactions, making it challenging to specify particular objects for editing~\cite{patashnik2023localizing,li2024zone,lin2024text}. Additionally, fully fine-tuning large models like SD~\cite{rombach2022highresolution,bar2022text2live,valevski2022unitune} is less feasible in research areas with limited computational resources.

To address the lack of facilities for training models on large-scale datasets, tuning-free methods have been developed to utilize pre-trained T2I SD~\cite{rombach2022highresolution}, referred to as \textit{zero-shot image editing}. These methods leverage a pre-trained T2I model with frozen weights to eliminate the need for adjusting the model's weights~\cite{Avrahami_2022_CVPR,meng2022sdedit,brack2024ledits++}. Most methods~\cite{hertz2022prompt,cao_2023_masactrl,liu2024towards,parmar2023zeroshot,Tumanyan_2023_CVPR} rely on attention mechanisms in SD models to preserve the original information of images, such as background and object identities. Specifically, some methods~\cite{Tumanyan_2023_CVPR,liu2024towards} swap or inject appropriate self-attention maps, while others, like~\cite{hertz2022prompt}, replace cross-attention maps to retain the content and structure of the original image during the synthesis process. However, these methods~\cite{Tumanyan_2023_CVPR,liu2024towards,titov2024guide} perform well when the edited object has a certain similarity to the original object in terms of shape, texture, and other attributes. Similarly, the approach in~\cite{hertz2022prompt} that replaces cross-attention maps requires the initial prompt and edited prompt to share similar words while incorporating different words. For instance, if the original sentence is `the photo of a red dog', the edited sentence might be `the photo of a yellow cat', where `red dog' and 'yellow cat' are the differing elements. In contrast, MasaCtrl~\cite{cao_2023_masactrl} adjusts self-attention to retain the current query features while replacing the key and value features. This approach ensures that the query features remain unchanged and are appropriately derived from the original semantic content guided by masks, rather than relying on rigidly swapped attention maps. As a result, MasaCtrl~\cite{cao_2023_masactrl} preserves the appearance of the original image in a non-rigid manner during synthesis. However, MasaCtrl~\cite{cao_2023_masactrl} controls the background and foreground simultaneously to obtain the semantic content of the corresponding original background and foreground at each appropriate step and layer. Thus, this approach lacks flexibility in controlling different image editing tasks; for example, we need the background to remain unchanged when the edited object resembles the original object. Additionally, a significant weakness of many methods is that, when editing images, they make changes to the overall image and cannot specifically edit individual objects within the image due to the condition of all image pixels and the text prompt in the cross-attention module (as shown in the middle images of Fig.~\ref{fig:cross_ex}). Some methods~\cite{hertz2022prompt,Avrahami_2022_CVPR,avrahami2023blended,couairon2022diffedit,li2024zone} address the challenge of local editing by blending the original latent noise with the edited noise, without considering the interaction between foreground and background, resulting in rigid editing. Subsequently, these methods lead to a substantial gap in addressing real image editing tasks.

Based on the existing extensive exploration of leveraging attention modules in SD to control the editing process and achieve desired outcomes, we analyze and clarify the semantic interaction of the components in attention and how to leverage them for real image editing (as detailed in Section~\ref{sec:preattn}). We then propose a novel zero-shot real image editing method, namely \textbf{C}ontext-\textbf{P}reserving \textbf{A}daptive \textbf{M}anipulation (CPAM), that leverages both self-attention and cross-attention. \blue{Importantly, CPAM can be directly applied to different diffusion backbones without architectural modification. In this work, we demonstrate its effectiveness across Stable Diffusion (SD) 1.5, SD2.1, and SDXL, showing that the proposed attention manipulation strategy generalizes well to various diffusion architectures.} CPAM excels in manipulating non-rigid objects, allowing for modifications to various aspects such as pose, view, or even specific objects or parts within the image. Importantly, CPAM retains the background and avoids modifications to unwanted objects or regions, thereby addressing issues faced by existing methods, enabling object removal or background replacement (Fig.~\ref{fig:overall}). Notably, these modifications occur without any model configuration or system architecture adjustments, eliminating the need for optimization or fine-tuning phases. Specifically, we introduce a preservation adaptation process that adjusts self-attention to independently control the object and background, effectively preserving the original objects' shapes, textures, and identities using masking techniques. Simultaneously, it ensures that the background remains undistorted or unwarped throughout the denoising process. Additionally, we propose the localized extraction module to avoid attention between the non-desired modified regions with the target prompt in the cross-attention. Therefore, our method enables localized editing, allowing for editing not only salient objects but also specific objects within the image. We propose different mask-guidance strategies to enable innovative image editing tasks by simply adjusting masks and enhancing regional manipulation by controlling object shapes. The source mask, representing the original object, and the target mask, controlling the edited outcome, are computed differently, allowing for flexible image editing.

In addition, we introduce a new \textbf{I}mage \textbf{M}anipulation \textbf{B}enchm\textbf{A}rk (IMBA) dataset, built upon TEdBench~\cite{kawar2023imagic}. We conduct a comprehensive user study on IMBA to assess the performance of our method against state-of-the-art text-guided image editing techniques utilizing SD. Extensive experiments and the user study unequivocally highlight the superiority of our proposed method, significantly outperforming state-of-the-art methods. \blue{The source code and data will be publicly released at the project page: \url{https://vdkhoi20.github.io/CPAM}.}

Our main contributions are as follows:
\begin{itemize}
    \item We introduce a novel tuning-free method, \textbf{C}ontext-\textbf{P}reserving \textbf{A}daptive \textbf{M}anipulation (CPAM), leveraging both self- and cross-attention for zero-shot real image editing.
    
    \item We design a preservation adaptation process to maintain object properties (\textit{e.g.}, pose, view, texture, identity, structure, color) while retaining the background.
    
    \item We develop a localized extraction module to avoid undesired prompt influence on unedited regions in cross-attention.
    
    \item We propose mask-guided strategies for flexible image manipulation and object shape tracking.
    
    \item We construct the IMBA benchmark dataset with enriched annotations for real image editing evaluation.
\end{itemize}

\section{Related work}
\label{sec:related}
\subsection{Image Manipulation Methods}

Several approaches required optimization or fine-tuning phases, which self-learned input images~\cite{mokady2022null,kawar2023imagic,NEURIPS2024_98b2b307,han2024proxedit}. DreamBooth~\cite{ruiz2023dreambooth} synthesized novel views of a given subject using 3–5 images of that subject and a target prompt. Textual Inversion~\cite{gal2022textual} optimized a new word embedding token for each concept. \blue{Recent works explore attention-based mechanisms to disentangle object and context representations for controllable image generation. MoA~\cite{wang2024moa} proposed mixture-of-attention for subject–context disentanglement, while KV-Mixture~\cite{parmar2025object} introduced object-level tokens for compositional generation.} Imagic~\cite{kawar2023imagic} generated novel poses and views by optimizing the target text embedding, fine-tuning model parameters, and interpolating between the approximate and target text embeddings. However, it struggled to maintain background consistency and realism, requiring careful optimization of embeddings for each prompt-image pair. MimicBrush~\cite{NEURIPS2024_98b2b307} trained on temporally consistent source-reference pairs extracted from large-scale video and image datasets, enabling reference-guided editing but requiring extensive training time due to the massive data volume. Null-Text Inversion (NULL)~\cite{mokady2022null} proposed optimal image-specific null-text embeddings for accurate reconstruction, combined with P2P~\cite{hertz2022prompt} techniques for real image editing. Pix2Pix~\cite{brooks2022instructpix2pix} performed full fine-tuning of the diffusion model by generating image-text-image triplets based on instructional input. However, the optimization and fine-tuning process is time-consuming and resource-intensive. Our method, instead, focuses on tuning-free techniques that eliminate the need for such processes.

\begin{figure*}[!t]
    \centering
    \includegraphics[width=\textwidth]{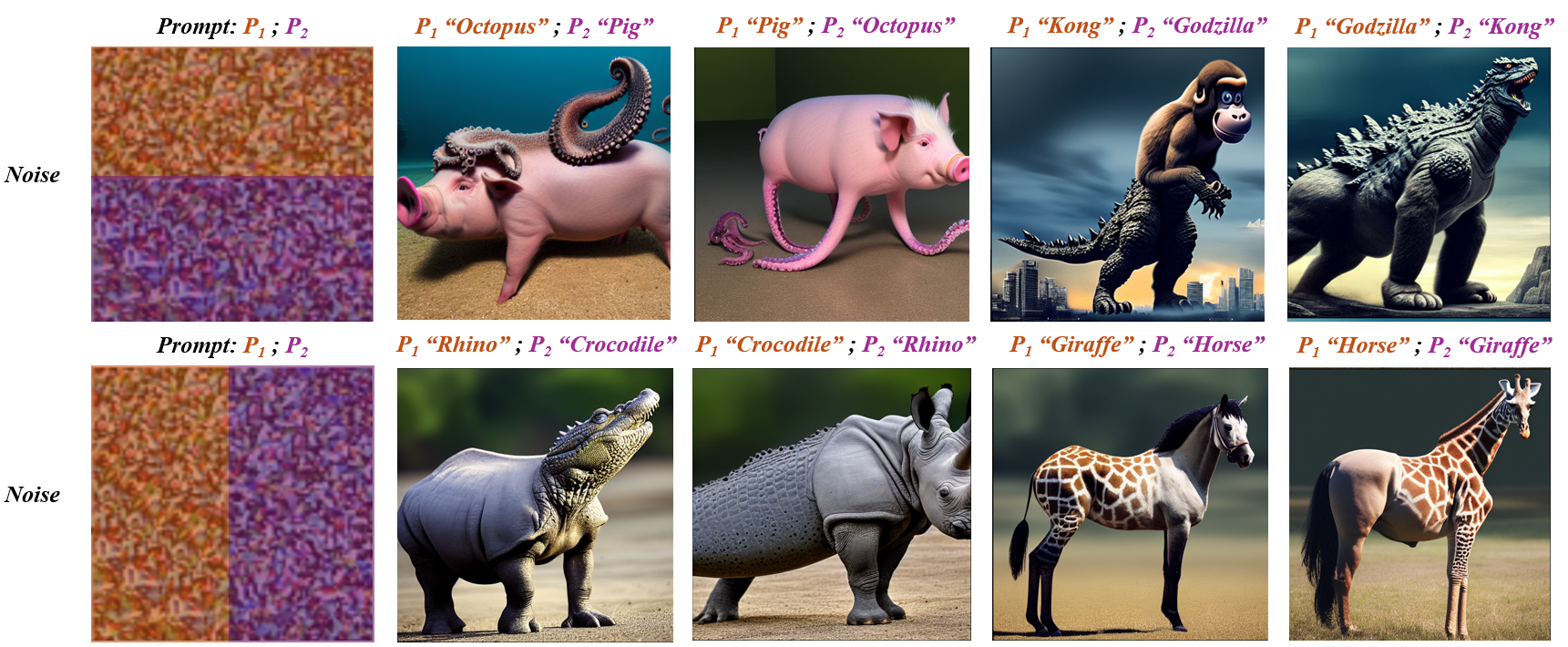}
    \vspace{-7mm}
    \caption{We perform multi-text guided synthesis, where each text prompt conditions a distinct part of the latent noise, effectively leading to results that exceed expectations and demonstrating the ability to condition each part with its respective prompt.} 
    \label{fig:insight}
    \vspace{-4mm}
\end{figure*}

\subsection{Zero-Shot Methods}
Zero-shot approaches focused on editing images directly during the denoising phase, eliminating the need for any fine-tuning or additional training. SDEdit~\cite{meng2022sdedit} introduced intermediate noise to an image, followed by denoising through a diffusion process conditioned on the desired edit. However, it exhibited a tradeoff between preserving the original image attributes and fully achieving the target text's intended changes. Blended Diffusion~\cite{Avrahami_2022_CVPR} facilitated local editing using gradient guidance based on the CLIP loss of the desired modified region and the target text prompt, without accounting for the interaction between foreground and background. However, blending this with the original image noise at each step led to rigid editing and inconsistency. Attend-and-Excite~\cite{chefer2023attendandexcite} generate images that fully convey the semantics of the given text prompt by progressively guiding the noised latent at each timestep, using the attention maps of the subject tokens from the prompt. LEDIT++~\cite{brack2024ledits++} proposed approaches for quickly and accurately inverting images and determining the appropriate direction for editing. Pix2Pix-Zero~\cite{parmar2023zeroshot} requires a large bank of diverse sentences from both source and target texts to form an edit direction. DDPM inversion~\cite{huberman2024edit} introduced an inversion method for DDPM, showing that the inversion maps encoded the image structure more effectively than the noise maps used in regular sampling, making them better suited for image editing. 

\blue{Image inpainting is a broader task that focuses on filling missing regions with plausible content, where object removal is a common scenario. While existing inpainting methods can remove objects, they are primarily designed for generic region completion rather than controllable real image editing. Additionally, most existing {zero-shot} diffusion editing methods are not specifically designed for object removal and often fail to completely remove the target object or introduce artifacts in surrounding regions. Our method instead explicitly separates object and background through mask-guided attention adaptation, enabling object removal and edits by simply specifying the target mask without requiring additional architectural modifications.}

\begin{figure*}[!t]
    \centering

    \subfloat[{\small CPAM pipeline.}]{
        \includegraphics[width=0.58\textwidth]{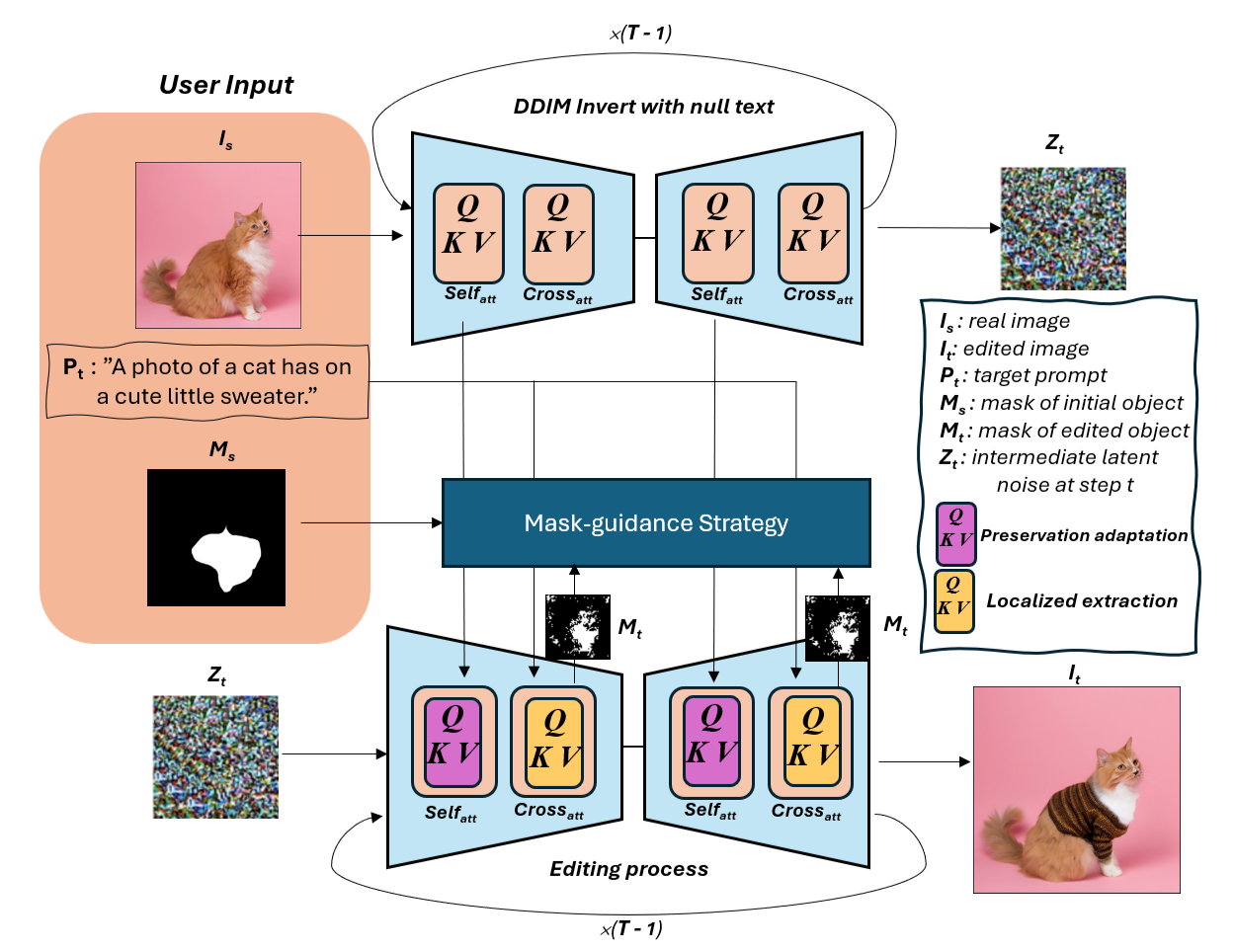}
    }
    \subfloat[{\small Preservation adaptation and localized extraction modules.}]{
        \includegraphics[width=0.4\textwidth]{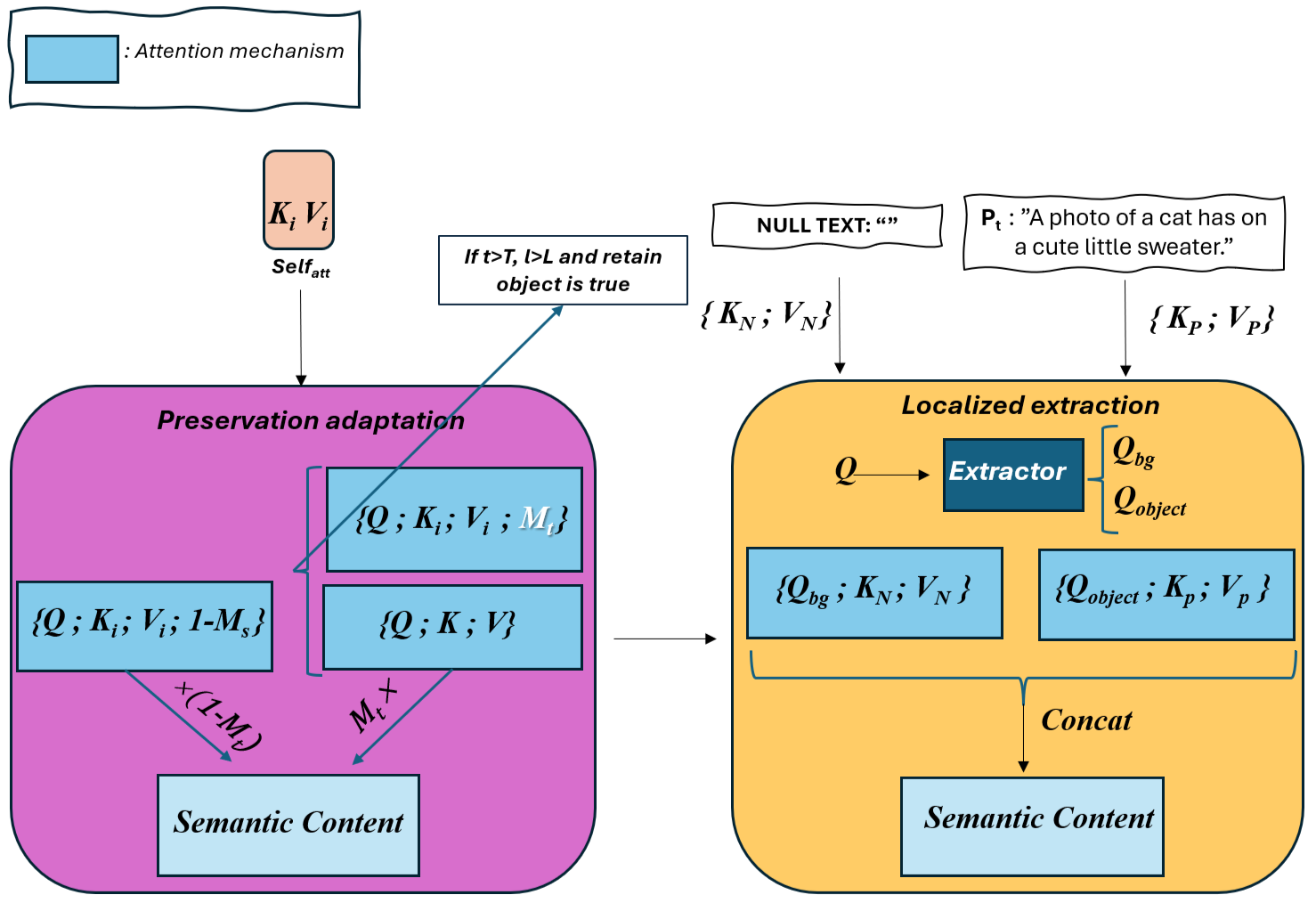}
    }
    \caption{\blue{Architecture of our proposed Context-Preserving Adaptive Manipulation (CPAM) for zero-shot real image editing. 
    Given a source image \(I_s\) and a target text prompt, the pipeline first obtains the source mask \(M_s\) through the \texttt{MaskInputModule}, which can derive masks from manual annotations, click-based interaction, or text prompts using SAM. 
    The source image is then inverted into latent noise using DDIM inversion with null-text guidance~\cite{song2020denoising}. 
    During the editing process, the source mask \(M_s\) defines the original object region and separates foreground from background, while the target mask \(M_t\) controls the desired edited structure and may evolve during diffusion, resulting the image \(I_t\). 
    The preservation adaptation (PA) module operates on self-attention to preserve background and object attributes outside the editing region, while the localized extraction (LE) module modulates cross-attention to restrict the influence of the target prompt to the specified editing area.}}
    \vspace{-3mm}
    \label{fig:overview}
\end{figure*}

\section{Preliminary Analysis of Attention Mechanism in Stable Diffusion}
\label{sec:preattn}
Within the Stable Diffusion (SD)~\cite{rombach2022highresolution}, the attention mechanism~\cite{46201} of the denoising U-Net, which includes both self-attention and cross-attention, is mathematically expressed as~$\text{Attention}(Q, K, V) = \text{Softmax}\left(\frac{QK^T}{\sqrt{d}}\right)V$, where \(Q\) represents the query features projected from spatial features, while \(K\) and \(V\) are the key and value features projected from spatial features (in self-attention layers) or textual embeddings (in cross-attention layers) using the respective projection matrices.

\subsection{Insights from cross-attention}
Cross-attention involves interactions between pixels and prompts (\textit{i.e.}, key and value features from textual embeddings). First, we observed that attending each prompt to different parts of latent noise allows each section to be conditioned by its respective prompt (as depicted in Fig.~\ref{fig:insight}). Second, null text does not affect the output, a phenomenon evident during training. Most diffusion models (DMs) utilize a classifier-free guidance~\cite{ho2021classifierfree}, randomly replacing text conditioning with null text at a fixed probability during training. As a result, when latent noise parts attend to null text, the corresponding pixels are perfectly reconstructed. Our method leverages this by directing attention to the pixels of specific objects using the text prompt, while background pixels attend to null text.

\subsection{Insights from self-attention}

Previous works~\cite{cao_2023_masactrl,liu2024towards,Tumanyan_2023_CVPR} show that self-attention features can be injected into U-Net layers for image translation, preserving semantic layout. Our key insight is that self-attention lets pixels connect with themselves, creating smooth transitions and consistent interactions. For example, in Fig.~\ref{fig:insight}, we apply two prompts (\textit{e.g.}, `crocodile' and `rhino') to two parts of latent noise, resulting in a cohesive and non-rigid outcome. Self-attention also helps each pixel determine which others to attend to, even when excluding a specific region, as shown in Fig.~\ref{fig:overall}, where all image pixels focus on the background pixels, excluding the teddy bear pixels, effectively removing it without disrupting the connections of the semantic in the image. 
By controlling self-attention, we minimize the impact on irrelevant areas while preserving image coherence.

\section{Proposed method}

\subsection{Overview}

    


Based on the insights in Section~\ref{sec:preattn}, we propose Context-Preserving Adaptive Manipulation (CPAM) to edit an image \( I_s \) using a source object mask \( M_s \) through the \texttt{MaskInputModule}, which can derive the mask in various ways, such as manual drawing, click-based extraction, or text prompts using SAM and a target text prompt $P_t$ to generate a new image $I_t$ that aligns with $P_t$. Notably, $I_t$ may spatially differ from $I_s$, modifying objects or background while keeping other regions unchanged. To achieve this, we introduce a preservation adaptation module that adjusts self-attention to align the semantic content from intermediate latent noise to the current edited noise, ensuring the retention of the original object and background during the editing process. To prevent unwanted changes from the target prompt in non-desired modified regions, we propose a localized extraction module that enables targeted editing while preserving the remaining details. Additionally, we propose mask-guidance strategies for diverse image manipulation tasks. The overall CPAM architecture is illustrated in Fig.~\ref{fig:overview}, and the zero-shot editing algorithm is outlined in Algorithm~\ref{algo:overview}.

\begin{figure*}[!t]
    \centering
    \includegraphics[width=\textwidth]{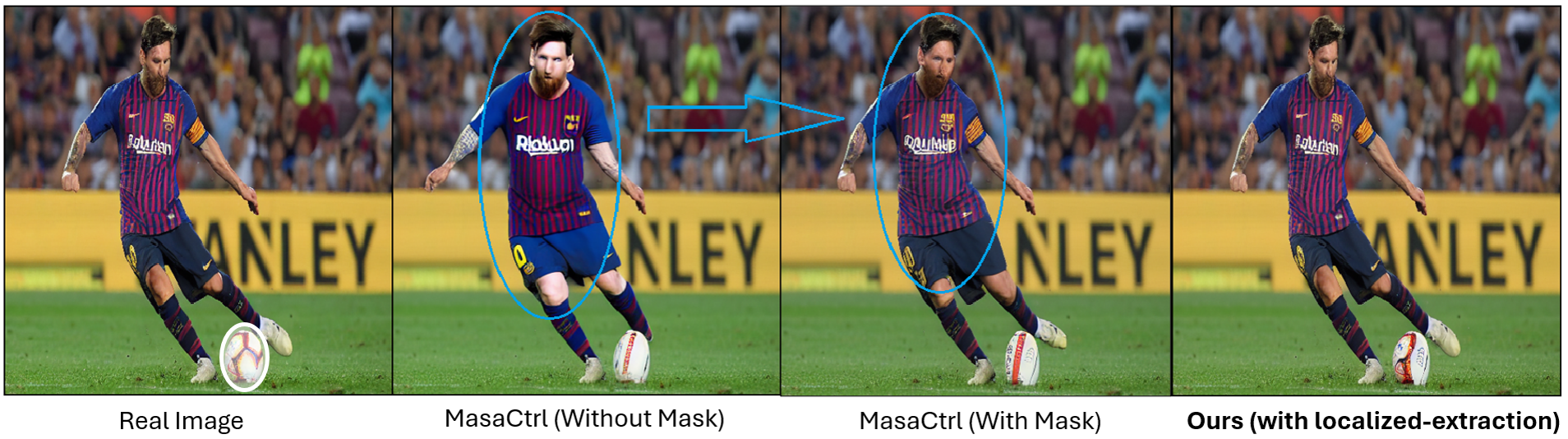}
    \vspace{-7mm}
    \caption{The real image (left), along with the edit prompt "\textit{Messi and a rugby}," and a mask suggests a desire to transform the soccer ball into a rugby ball. In two middle images, MasaCtrl~\cite{cao_2023_masactrl} produces another instance of Messi without mask guidance and retains the background with mask guidance, but the problem remains unsolved (marked by blue annotation). On the other hand, the localized extraction in our CPAM can successfully preserve the background while transforming the soccer ball into the rugby ball (the right image).}
    \label{fig:cross_ex}
    \vspace{-5mm}
\end{figure*}

\begin{figure}[!t]
\vspace{-2mm}
\begin{algorithm}[H]
\small
\caption{Zero-Shot Real Image Editing}\label{algo:overview}
\textbf{Inputs:} 
A target prompt \(P_t\), a source mask \(M_s\), the intermediate latent noises \(z_i\), the target initial latent noise map \(z_T\). \\
\textbf{Output:} Edited latent map \(z_0\).

\begin{enumerate}
    \item \textbf{For} \(t = T, T-1, \ldots, 1\) \textbf{do:}
    \begin{enumerate}
        \item \(M_t \leftarrow \text{Mask-guidance strategy}(\text{cross-attention maps}, M_s)\)
        \item \(\{\_,K_i, V_i\} \leftarrow \epsilon_\theta(z_i, t)\)
        \item \(\{Q, K, V\} \leftarrow \epsilon_\theta(z_t, t)\)
        \item \(\text{inputs} \leftarrow (Q,K,V,K_i,V_i,M_s,M_t) \)
        \item \(\text{self-attention} \overset{\text{adapt}}{\longleftarrow} \text{Preserving-adaptation}(\text{inputs})\)
        \item \(\text{cross-attention} \overset{\text{inject}}{\longleftarrow} \text{Localized-extraction} (Q,P_t,P_{null},M_t)\)
        \item \(\epsilon \leftarrow \epsilon_\theta(z_t, P_t, t,\text{self-attention},\text{cross-attention})\)
        \item \(z_{t-1} \leftarrow \text{Sample}(z_t, \epsilon)\)
      
    \end{enumerate}
    \item \textbf{End For.}
\end{enumerate}
\textbf{Return:} \(z_0\).
\end{algorithm}
\vspace{-5mm}
\end{figure}

\subsection{Preservation Adaptation}
\label{sec:preseradap}

In this section, we describe the self-attention adaptation process, which preserves the original image's appearance by independently adapting the semantic content from intermediate latent noise to the edited image.

\subsubsection{Background preservation adaptation}
To adapt semantic content from intermediate latent noise during the denoising step \(t\), we retain the query features \(Q\) and extract the original semantic content from the key and value features \(K\) and \(V\) at that self-attention layer. We then apply attention guided by the mask \(M_s\). The semantic content of the background \(SC_{bg}\) can be formulated as:
\begin{equation}
    \label{eq:bgpreser}
    \text{SC}_\text{bg} = \text{Att}(Q, K_i, V_i; 1-M_s),    
\end{equation}
where \(K_i\) and \(V_i\) correspond to the key and value features of the intermediate latent noise, respectively, and Att is the attention mechanism.

\subsubsection{Object preservation adaptation}
 
Preserving the object's semantic content is more difficult than maintaining the background because it requires adapting the original object's features to fit a new shape, pose, or view. We carefully control this adaptation after \( S \) step and layer \( L \), while generating new shapes, poses, or views. Thus, the semantic content of the object (foreground) \( SC_{fg} \) can be formulated as follows:
\begin{equation}
    \label{eq:fgpreser} 
    \text{SC}_\text{fg} = 
    \begin{cases}
        \text{Att}(Q, K_i, V_i;M_t), \text{ If } t > T \text{ and } l > L \\ \text{ and the object is retained},
        \\
        \text{Att}(Q, K, V), \text{ otherwise,}
    \end{cases}
\end{equation}
where $t$, $l$, $T=3$, $L=8$ denote step and layer, respectively. \(V_{i}\) are the value feature of intermediate latent noise at step $i$. \(K\) and \(V\) are the key and value features of current noise, respectively. \(M_t\) is the target mask of the edited object, and Att is the attention mechanism.

\subsubsection{Location adaptation}

The aim of this module is to provide precise control over the foreground and background during image editing, allowing for independent adjustments. By separately deriving the semantic content of the background from Equation.~\ref{eq:bgpreser} and the foreground from Equation.~\ref{eq:fgpreser}, and aligning them with the target mask \(M_t\), we achieve flexible, region-specific edits. This process ensures that modifications occur only in designated areas, keeping the rest of the image intact. Consequently, the combined semantic content \(SC\), guided by the target mask, is formulated as follows:
\begin{equation}
\label{eq:location}
SC = M_t \odot SC_{\text{foreground}} + (1 - M_t) \odot SC_{\text{background}},
\end{equation}
where \(\odot\) denotes the element-wise multiplication. However, applying self-attention independently to the foreground and background leads to a lack of interaction, resulting in rigid editing. To maintain overall image coherence, we randomly apply normal self-attention in 10\% of the layers. This approach minimizes unintended distortions and yields more natural results during synthesis.

\subsection{Localized Extraction}
\label{sec:localextrac}
Despite significant efforts to adapt the content of the original image in our preservation adaptation module, the non-desired modified spatial region may appear distorted (as shown in the third image in Fig.~\ref{fig:cross_ex}). This distortion arises because all pixels of the image attend to tokens of the text prompt, affecting both the foreground (\textit{i.e.}, object) and background, including non-desired modified regions. To address this issue, we introduce a localized extraction mechanism, allowing for editing only a specific object without distorting the rest. This mechanism applies attention to the extracted object’s spatial pixels from the feature query to the target prompt, while the remaining pixels attend to the null text prompt:
\begin{equation}
\begin{split}
\text{LE}(Q, K, V) = 
&\ \text{Att} (\text{Extract}(Q, M_t), K_t, V_t) \\
&\oplus \text{Att} (\text{Extract}(Q, 1 - M_t), \text{Null}(K), \text{Null}(V)),
\end{split}
\end{equation}
where $\oplus$ represents concatenation, \(\mathrm{Extract}(Q,M)\) extracts the object's spatial pixels from the feature query $Q$ where $M = 1$, and Att is the attention mechanism.

\subsection{Mask-Guidance Strategy}
\label{sec:maskref}

Achieving both rigid and non-rigid semantic changes within a unified framework for diverse image manipulation tasks is a notable challenge. Our method is designed to simplify this by requiring only adjustments to mask settings. We present various mask guidance strategies tailored to different editing needs. The source mask \(M_s\) refers to the mask of the object or region the user wishes to edit, which can be provided through various methods such as manual drawing, extraction from clicks, or text prompts using SAM~\cite{kirillov2023segment}. Meanwhile, the target mask \(M_t\) can be obtained as follows:

\begin{itemize}

    \item Replacing an object or changing the object pose, view: \(M_t\) is achieved by aggregating cross-attention maps across all steps and layers or extending the convex hull of \(M_s\).
    \item Altering background: When \(M_s\) is the mask of the background, we assign \(M_t = M_s\).
    \item Removing object: We assign \(M_t = 0\), meaning in our preservation adaptation module only adapting the semantic content of the original background (refer to Equation~\ref{eq:location}).
    \item Modifying a specific spatial region (\textit{e.g.}, adding items): We simply assign \(M_t = M_s\) or slightly expand \(M_s\).    
\end{itemize}

\textbf{Mask refinement. } When manipulating an object, its shape may change during diffusion steps. To address this, we refine the mask according to the target prompt during the denoising process. The target mask \(M_t\) is automatically obtained by aggregating cross-attention maps. Initially, for the first \(T_m\) steps, we use the source mask \(M_s\), then transition to \(M_t\), which can be cloned from \(M_s\) or derived from the generation process. Additionally, we avoid closely segmenting both the target and original objects to prevent overly rigid editing and the leakage of underlying shape information.

\subsection{Discussion on Differences with Prior Methods}
\begin{figure}[!t]
    \centering
    \includegraphics[width=\linewidth]{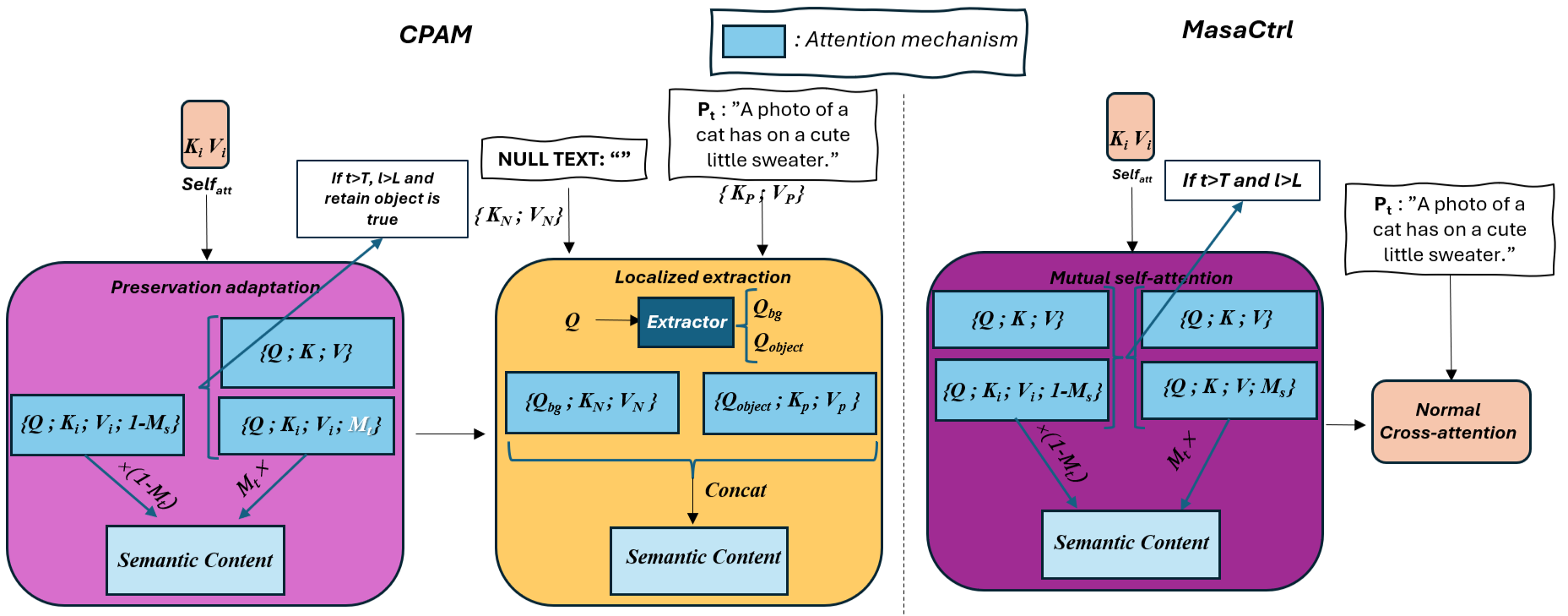}
    \caption{\blue{Comparative details of our CPAM and MasaCtrl~\cite{cao_2023_masactrl}.}}
    \label{fig:compare_cpam_masa}
    \vspace{-5mm}
\end{figure}

\blue{To clarify the conceptual differences between CPAM and closely related diffusion editing approaches, we compare our method with representative attention-control methods including MasaCtrl~\cite{cao_2023_masactrl}, PnP~\cite{Tumanyan_2023_CVPR}, and DiffEdit~\cite{couairon2022diffedit}. Although these approaches also manipulate internal diffusion representations, they differ from CPAM in several fundamental aspects.}

\blue{MasaCtrl~\cite{cao_2023_masactrl} is the method most closely related to our proposed approach. MasaCtrl manipulates self-attention by jointly querying semantic information from the source image across layers and diffusion steps in order to preserve structural consistency. However, object and background information are propagated together during the denoising process. In contrast, CPAM introduces mask-guided attention adaptation that explicitly separates object and background queries, enabling independent control of edited and preserved regions (Fig.~\ref{fig:compare_cpam_masa}).}

\blue{PnP~\cite{Tumanyan_2023_CVPR} preserves image structure by directly injecting intermediate spatial features and attention maps extracted from the source image. While this strategy effectively maintains the global layout, it propagates object and background features jointly without explicit spatial disentanglement. As a result, the editing process remains globally coupled and provides limited control over object-level manipulation.} \blue{Meanwhile, DiffEdit~\cite{couairon2022diffedit} localizes editing regions by estimating masks from noise prediction differences between the source and target prompts. After mask estimation, the method performs editing by aligning intermediate noise representations during the denoising process. This design often leads to rigid edits and limited flexibility when handling non-rigid transformations such as pose or view changes.}

\blue{In contrast, CPAM explicitly disentangles object and background representations via mask-guided attention adaptation and further introduces a localized extraction mechanism to restrict prompt influence to target regions. Together, these components enable flexible object-level editing while preserving background consistency, yielding more precise and controllable results than prior attention-control methods.}

\section{Experiments}

\subsection{Implementation Details}
\label{sec:experimental_details}

All experiments were conducted on a machine with a single A100 GPU. \blue{Our proposed CPAM was employed using the publicly available SD-1.5, SD-2.1 and SDXL models.} We initially encode the image into latent code by variational autoencoder~\cite{kingma2014auto} and invert to noise using the deterministic inversion technique of DDIM~\cite{song2020denoising} with null-text guided. We employed guidance scale 7.5, 50 inference steps.

\subsection{Image Manipulation BenchmArk (IMBA) Dataset} 
\label{imba}

\begin{figure}[t!]
    \centering
    \includegraphics[width=\linewidth]{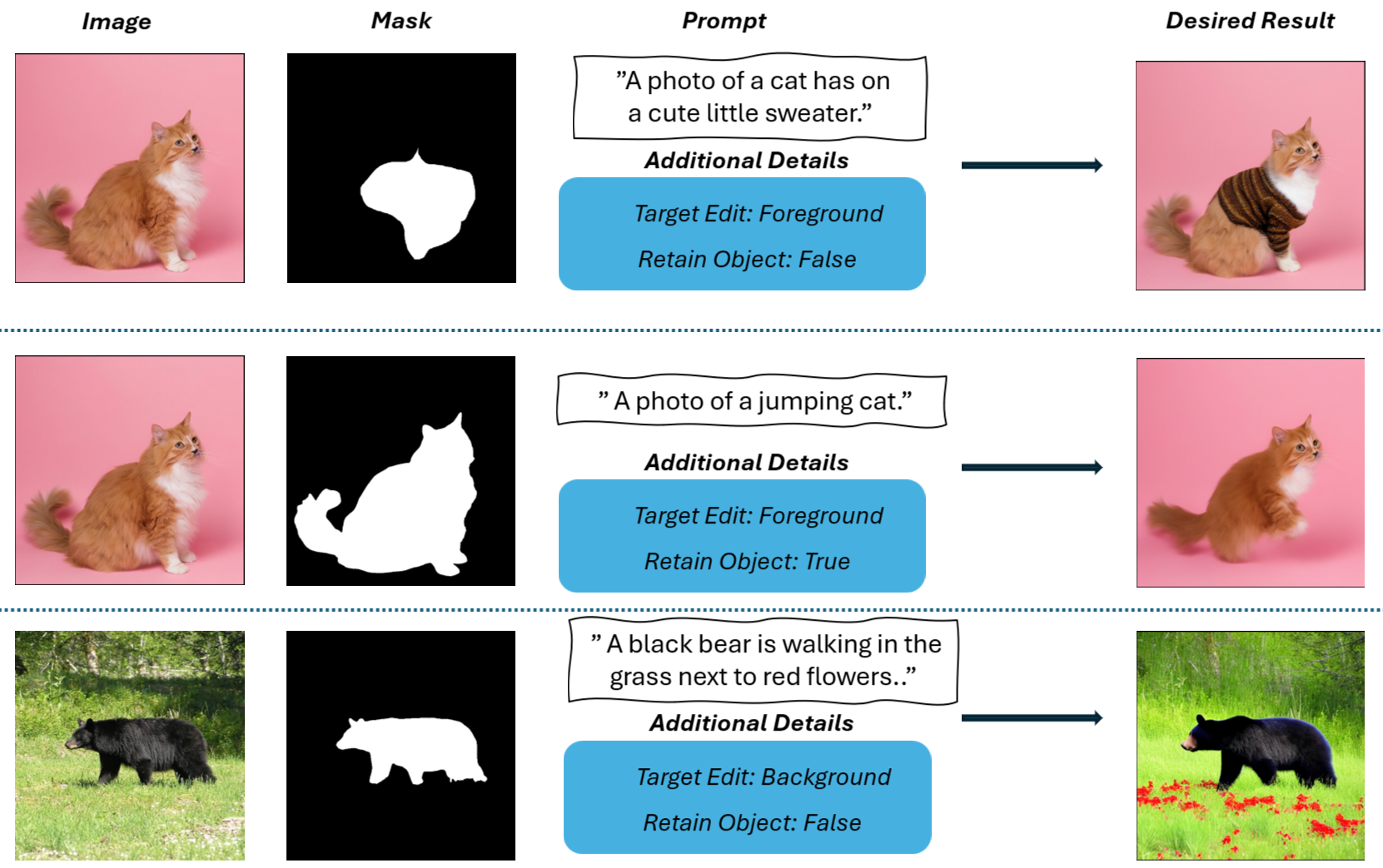}
    \vspace{-5mm}
    \caption{\blue{Visualization of Image Manipulation BenchmArk (IMBA).}}
    \label{fig:IMBA}
    \vspace{-5mm}
\end{figure}

\blue{Textual Editing Benchmark (TEdBench)~\cite{kawar2023imagic} is a widely adopted benchmark for evaluating non-rigid real image editing, consisting of 100 pairs of input images and target textual edits. Many prior works evaluate their methods either on the full TEdBench dataset or on subsets of its samples~\cite{cao_2023_masactrl,hertz2022prompt,kawar2023imagic,brack2024ledits++}. However, TEdBench primarily provides image–text editing pairs and does not include detailed annotations describing user editing preferences or explicit object-level editing constraints.}

\blue{To address these limitations, we introduce Image Manipulation BenchmArk (IMBA), which extends TEdBench with richer annotations designed for controllable object-level image editing. As illustrated in Fig.~\ref{fig:IMBA}, IMBA augments each sample with additional information such as alteration masks, object prompts, and explicit editing preference labels, enabling more precise control and evaluation of object-level manipulation. In particular, the dataset distinguishes between different editing intents, including object retention, object modification, and background alteration, allowing us to evaluate how well methods preserve scene structure while editing specific objects. \bluex{More importantly, IMBA emphasizes localized editing scenarios, where modifications are intended to affect only a specific object or region while preserving the remaining image content. Furthermore, IMBA includes scenes with multiple objects, creating ambiguous cases in which textual instructions alone may be insufficient to uniquely identify the desired editing target}. In total, IMBA contains 104 editing samples, including 43 cases requiring object retention, 97 cases involving object modification, and 7 cases involving background alteration. These annotations provide a more structured and controlled evaluation setting compared to the original TEdBench.}

\subsection{Evaluation Metrics}

\blue{We evaluate text-image alignment using CLIPScore and assess background preservation using LPIPS, DreamSim (DS)~\cite{fu2023dreamsim}, and RMSE measured on the ground-truth background mask. For LPIPS, we follow the official implementation using the AlexNet backbone~\cite{zhang2018unreasonable}, which is commonly adopted for perceptual similarity evaluation in prior diffusion editing works~\cite{couairon2022diffedit,Tumanyan_2023_CVPR,brack2024ledits++,karras2020analyzing}. While LPIPS measures low-level appearance differences based on CNN features, DreamSim~\cite{fu2023dreamsim} has been shown to better correlate with human perceptual judgments. Therefore, we report both metrics together with the standard RMSE to provide a comprehensive evaluation of editing quality.}

\subsection{Comparison with State-of-the-Arts}

\subsubsection{Compared Methods}

\blue{We compare our proposed CPAM with several state-of-the-art real image editing approaches built upon Stable Diffusion, including P2P+NULL~\cite{hertz2022prompt,mokady2022null}, SDEdit~\cite{meng2022sdedit}, MasaCtrl~\cite{cao_2023_masactrl}, PnP~\cite{Tumanyan_2023_CVPR}, FPE~\cite{liu2024towards}, DiffEdit~\cite{couairon2022diffedit}, Pix2Pix-Zero~\cite{parmar2023zeroshot}, LEDITS++~\cite{brack2024ledits++}, and the fine-tuning method Imagic~\cite{kawar2023imagic}. These approaches represent different editing paradigms, including prompt-based editing, attention manipulation, feature injection, and model fine-tuning.} For all methods, we utilized publicly available official code, with the exception of Imagic~\cite{kawar2023imagic}. For Imagic, we evaluated publicly available results and leveraged community-developed code from the Diffusers library on Hugging Face.

\blue{We performed a grid search across the hyperparameter ranges specified for each method while keeping other parameters at their default settings. Initially, a wider range of values was explored to define reasonable boundaries, after which edge values that resulted in performance declines were discarded.}
\blue{
\begin{itemize}
    \item P2P~\cite{hertz2022prompt} + NULL~\cite{mokady2022null}: The initial prompt is obtained using CLIP~\cite{radford2021learning}, with the self-attention replacement ratio set between 0.4 and 0.7.
    \item SDEdit~\cite{meng2022sdedit}: The number of diffusion steps ranges from 25 (corresponding to strength 0.5 with 50 total steps) to 40 (strength 0.8 with the default 50 steps).
    \item MasaCtrl~\cite{cao_2023_masactrl}: The step query is set to 4, and the layer query ranges from 10 to 14. Three mask configurations are used: no mask guidance, explicit ground-truth mask, and automatically aggregated mask.
    \item PnP~\cite{Tumanyan_2023_CVPR}: The self-attention replacement and feature injection ratios are set between 0.5 and 0.8, with 50 inference steps.
    \item FPE~\cite{liu2024towards}: The self-attention replacement ratio is set between 0.5 and 0.8.
    \item DiffEdit~\cite{couairon2022diffedit}: The initial prompt is obtained using CLIP~\cite{radford2021learning}, with an explicit ground-truth mask.
    \item Pix2Pix-Zero~\cite{parmar2023zeroshot}: Five source prompts and five target prompts are generated using the Flan-T5-XL model~\cite{chung2024scaling}.
    \item LEDIT++~\cite{brack2024ledits++}: Uses 50 inversion steps, with a skip ratio between 0.1 and 0.3, and supports both explicit ground-truth masks and automatically generated masks.
    \item Imagic~\cite{kawar2023imagic}: Uses 500 steps for text embedding optimization and 1000 steps for model fine-tuning, with $\alpha$ ranging from 0.1 to 2.0.
\end{itemize}
}

\begin{figure*}[!t]
    \centering
    \includegraphics[width=\textwidth]{images/qualitative_visualization.jpg}
    \vspace{-5mm}
    \caption{ \blue{Qualitative comparison of our proposed CPAM with state-of-the-art methods across multiple real image editing tasks. Results are shown on different diffusion backbones, including SD1.5, SD2.1, and SDXL. CPAM consistently produces more coherent edits while preserving background structure and scene layout. Please zoom in for clearer details.}}
    \label{fig:qualitative}
    \vspace{-5mm}
\end{figure*}

\subsubsection{Qualitatitve Evaluation}

\begin{table}[!t]
\centering
\caption{ Functional comparison across editing methods. \textbf{Local Edit} denotes region-specific editing. \textbf{Obj. Removal} refers to the ability to remove objects.  \textbf{Caption-Free} indicates the method does not require the original image caption. \textbf{Mask Ctrl} refers to mask-based region control. \textbf{Hi-Guidance} indicates compatibility with high classifier-free guidance scales. \textbf{FT} indicates fine-tuning method.}
\label{tab:functional-capabilities}
\vspace{-1mm}
\resizebox{\linewidth}{!}{
\begin{tabular}{lccccc}
\toprule
\textbf{Method} & 
\textbf{Local Edit} & 
\textbf{Obj. Removal} & 
\textbf{Caption-Free} & 
\textbf{Mask Ctrl} & 
\textbf{Hi-Guidance} \\
\hline
SDEdit~\cite{meng2022sdedit} & \xmark & \xmark & \cmark & \xmark & \xmark \\
MasaCtrl~\cite{cao_2023_masactrl} & \cmark & \xmark & \cmark & \cmark & \xmark \\
PnP~\cite{Tumanyan_2023_CVPR} & \xmark & \xmark & \xmark & \xmark & \xmark \\
FPE~\cite{liu2024towards} & \xmark & \xmark & \xmark & \xmark & \xmark \\
DiffEdit~\cite{couairon2022diffedit} & \cmark & \xmark & \cmark & \cmark & \xmark \\
Pix2Pix-Zero~\cite{parmar2023zeroshot} & \xmark & \xmark & \xmark & \xmark & \xmark \\
LEDITS++~\cite{brack2024ledits++} & \cmark & \xmark & \cmark & \cmark & \xmark \\
Imagic~\cite{kawar2023imagic} \textbf{(FT)} & \xmark & \xmark & \xmark & \xmark & \xmark \\
\rowcolor{lightgray} \textbf{CPAM (Ours)} & \cmark & \cmark & \cmark & \cmark & \cmark \\
\bottomrule
\end{tabular}
}
\vspace{-4mm}
\end{table}

\blue{As summarized in Table~\ref{tab:functional-capabilities}, CPAM is the only method that simultaneously supports all key functional capabilities for real image editing, including localized editing, object removal, and editing without requiring the original image caption. This highlights the flexibility of CPAM in handling diverse editing scenarios and distinguishes it from both zero-shot and fine-tuning-based methods.}

\blue{To further evaluate the robustness and generalization capability of our framework, we extend CPAM to more advanced diffusion backbones, including SD1.5, SD2.1, and SDXL. Integrating CPAM into different architectures just requires adapting to different attention layer configurations and feature representations. Nevertheless, the proposed framework can be applied without architectural modifications, demonstrating its compatibility with modern diffusion models.}

\blue{Figure~\ref{fig:qualitative} presents qualitative comparisons between CPAM and existing approaches across multiple real image editing tasks. Our method consistently produces more coherent edits while preserving background structure and scene layout. In contrast, many existing methods introduce unintended changes outside the target editing region due to globally coupled editing operations. CPAM effectively modifies diverse image attributes such as pose, viewpoint, background context, and object appearance while maintaining consistent background details.}

\begin{table}[t!] 
\centering 
\caption{\blue{Comparison with state-of-the-art methods using CLIPScore to measure text-image alignment, LPIPS, DreamSim (DS), and RMSE to evaluate background preservation.} \textcolor{red}{\textbf{Bold}}, \textcolor{blue}{\underline{underline}}, and \textcolor{olive}{\textit{italic}} indicate the best, second best, and third best scores, respectively. Imagic~\cite{kawar2023imagic} is the fine-tuning method (\textbf{FT}).}

\label{tab:quantitative} 
\begin{tabular}{lcccc} 
\toprule 
\textbf{Method} & \textbf{CLIP} $\uparrow$ & \textbf{LPIPS} $\downarrow$ & \textbf{DS} $\downarrow$ & \textbf{RMSE} $\downarrow$ \\ 
\hline 
SDEdit~\cite{meng2022sdedit} & 28.19 & 0.386 & 0.239 & 57.16 \\ 
MasaCtrl~\cite{cao_2023_masactrl} & 28.82 & 0.246 & 0.121 & 32.51 \\ 
PnP~\cite{Tumanyan_2023_CVPR} & 29.03 & 0.238 & 0.075 & 24.36 \\ 
FPE~\cite{liu2024towards} & 29.02 & 0.201 & \textcolor{blue}{\underline{0.056}} & \textcolor{olive}{\textit{22.06}} \\ 
DiffEdit~\cite{couairon2022diffedit} & 28.58 & 0.182 & 0.080 & 35.81 \\ 
Pix2Pix-Zero~\cite{parmar2023zeroshot} & 27.01 & 0.229 & 0.117 & 31.30 \\ 
LEDITS++~\cite{brack2024ledits++} & 28.74 & 0.210 & 0.104 & 43.30 \\ 
Imagic~\cite{kawar2023imagic} (\textbf{FT}) & \textcolor{red}{\textbf{30.34}} & 0.462 & 0.286 & 81.55 \\ 
\rowcolor{lightgray} \textbf{CPAM-SD1.5 (Ours)} & \textcolor{olive}{\textit{29.26}} & \textcolor{olive}{\textit{0.180}} & \textcolor{olive}{\textit{0.072}} & 23.42 \\ 
\rowcolor{lightgray} \textbf{CPAM-SD2.1 (Ours)} & 29.08 & \textcolor{blue}{\underline{0.125}} & \textcolor{red}{\textbf{0.044}} & \textcolor{blue}{\underline{19.13}} \\ 
\rowcolor{lightgray} \textbf{CPAM-SDXL (Ours)} & \textcolor{blue}{\underline{29.77}} & \textcolor{red}{\textbf{0.118}} & \textcolor{red}{\textbf{0.044}} & \textcolor{red}{\textbf{18.90}} \\ 

\bottomrule 
\end{tabular} 
\vspace{-6mm}
\end{table}

\subsection{Quantitative Evaluation}

\blue{Quantitative results are summarized in Table~\ref{tab:quantitative}.} \blue{We excluded the evaluation of P2P~\cite{hertz2022prompt} combined with Null text inversion technique\cite{mokady2022null} due to its reliance on an initial prompt that often leads to unchanged outputs. Results reveal a clear trend: methods like SDEdit~\cite{meng2022sdedit} often yield unrealistic results due to their dependence on noise strength parameters, which can disrupt semantic consistency. MasaCtrl~\cite{cao_2023_masactrl} lacks precise control over background and foreground elements during denoising, leading to unwanted alterations. PnP~\cite{Tumanyan_2023_CVPR} preserves the background but often fails to meet the target prompt, while FPE~\cite{liu2024towards} generates minimal visible changes due to its high reliance on self-attention maps. Pix2Pix-Zero~\cite{parmar2023zeroshot} struggles in real image editing tasks due to its dependence on closely matched prompts. Additionally, DiffEdit~\cite{couairon2022diffedit} and LEDIT++~\cite{brack2024ledits++} often capture the entire object when generating masks based on noise estimation, resulting in unwanted modifications. Although Imagic~\cite{kawar2023imagic} excels in user satisfaction, it frequently struggles with background retention and can produce misalignments or unwanted alterations, and requires more time consuming for fine-tuning and optimizing for each image prompt pair. In contrast, our CPAM demonstrates a more robust performance in preserving both object integrity and background details, effectively executing complex edits without sacrificing quality. This combination of qualitative and quantitative evaluations underscores the effectiveness of our approach in the context of modern image editing techniques.}

\blue{Additionally, CPAM consistently achieves strong performance across different diffusion backbones. In particular, CPAM-SDXL obtains the best background preservation performance with {LPIPS of 0.118} and {RMSE of 18.90}, while maintaining competitive text-image alignment with {CLIPScore of 29.77}. CPAM-SD2.1 also achieves the best DreamSim score and the second-best RMSE, demonstrating the effectiveness of the proposed method across different model architectures. We observe that the performance of CPAM further improves when integrated with stronger diffusion backbones such as SD2.1 and SDXL. This behavior is consistent with the architectural characteristics of these models. For example, SDXL contains significantly more attention layers than SD1.5, providing richer attention interactions that benefit our attention manipulation strategy, while SD2.1 introduces improved text conditioning and feature representations. Since CPAM directly operates on self- and cross-attention mechanisms, it naturally benefits from these stronger architectural capacities.}




\subsection{Ablation Study}

\subsubsection{Effectiveness of Proposed Components}

\begin{figure}[t!]

    \centering
    \includegraphics[width=\linewidth]{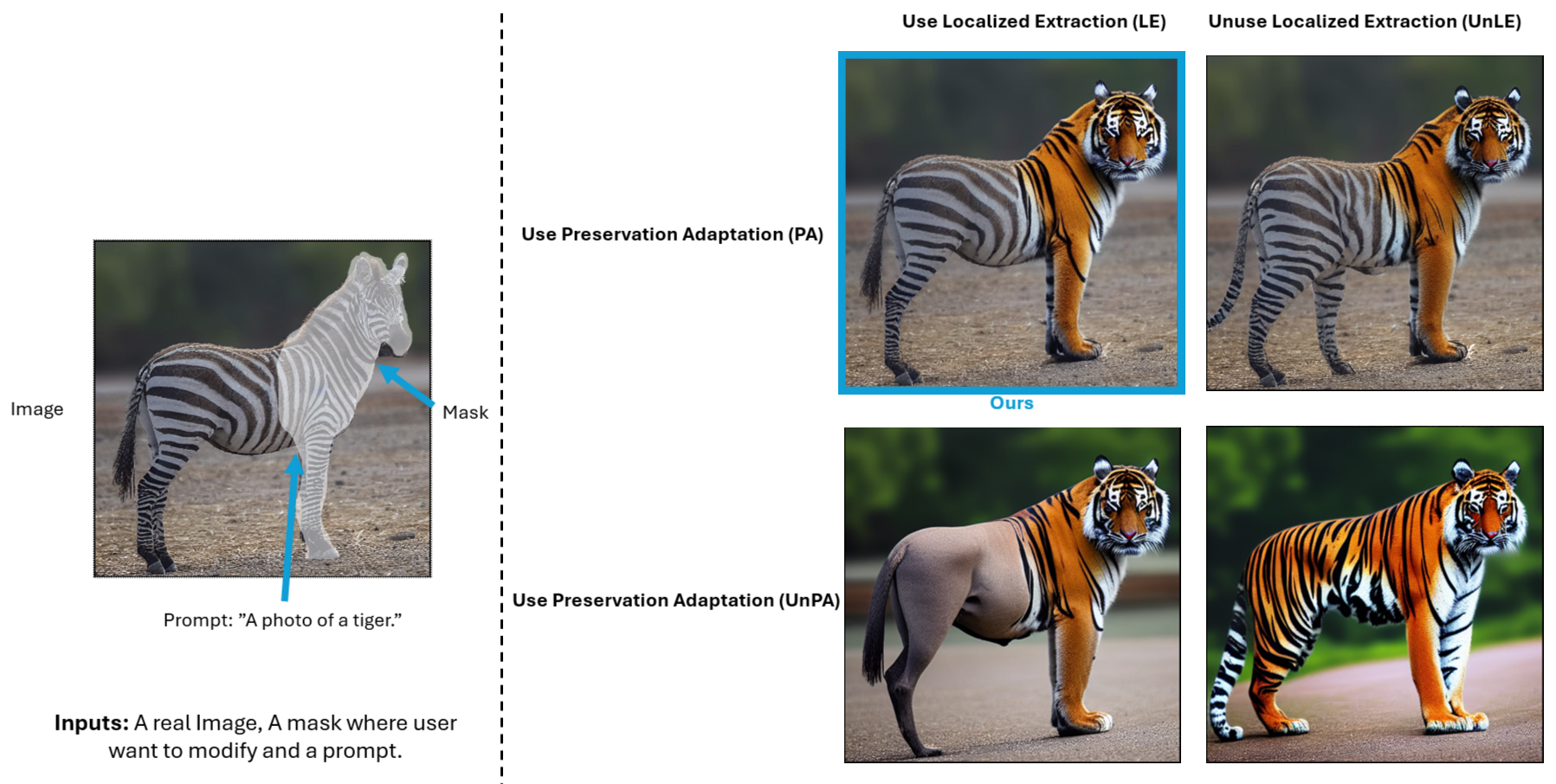}
    \vspace{-6mm}
    \caption{Effectiveness of PA and LE modules. Enabling both PA and LE localizes edits to prompt-specified regions, leaving non-masked areas intact.}
    \label{fig:isolate_module}
\end{figure}

\blue{We assess our proposed modules’ effectiveness using the Preservation Adaptation (PA) to query and preserve original semantic content in non-masked regions, while the Localized Extraction (LE) applies null text conditioning to maintain their attributes, preventing text-prompt-induced distortions. Figure~\ref{fig:isolate_module} shows that enabling both PA and LE results in edits localized to the specified regions defined by the text prompt, while the non-masked regions remain unaffected. Without using LE, the horse’s left side retains its color, but the leg and tail morph into tiger-like features per the prompt. Disabling PA alters the color while preserving the shape of legs and tail, resembling a zebra. Disabling both generates a full tiger, adhering to the prompt without retaining original semantics. }

\blue{To further quantify the contribution of each component,
we conduct an ablation study on the SDXL backbone (Table~\ref{tab:ablation}). Removing either module significantly degrades background preservation performance. In particular, removing LE increases LPIPS from 0.118 to 0.204 and RMSE from 18.90 to 33.26, indicating that the influence of the text prompt spreads to pixels across the entire image instead of being confined to the target editing region. Consequently, the editing process becomes less localized and leads to undesired background alterations. When PA is removed, the degradation becomes even more severe, with RMSE increasing to 38.33 and DreamSim rising to 0.181. This behavior occurs because PA is responsible for preserving background semantic features during the diffusion process. Interestingly, the CLIPScore slightly increases when PA is removed, as the editing process becomes less constrained and the generated image can more freely align with the target text prompt. However, this comes at the cost of substantial background distortion and the loss of texture details and object identities. Overall, these results demonstrate that both LE and PA are crucial for balancing accurate text alignment with stable background preservation, with PA playing a particularly important role in maintaining scene consistency.}

\begin{table}[t!] 
\centering 
\caption{\blue{Ablation study by removing Localized Extraction (LE) and Preservation Adaptation (PA) modules. Removing these components significantly degrades background preservation, demonstrating their importance in maintaining scene consistency during editing. The best scores are shown in \textcolor{red}{\textbf{Bold}}.}}
\label{tab:ablation}
\vspace{-2mm}
\begin{tabular}{lcccc} 
\toprule 
\textbf{Method} & \textbf{CLIP} $\uparrow$ & \textbf{LPIPS} $\downarrow$ & \textbf{DS} $\downarrow$ & \textbf{RMSE} $\downarrow$ \\ 
\hline 

\rowcolor{lightgray} \textbf{CPAM-SDXL (Ours)} & 29.77 & \textcolor{red}{\textbf{0.118}} & \textcolor{red}{\textbf{0.044}} & \textcolor{red}{\textbf{18.90}} \\ 

\textbf{CPAM-SDXL w/o LE} & 28.39 & 0.204 & 0.123 & 33.26 \\ 
\textbf{CPAM-SDXL w/o PA} & \textcolor{red}{\textbf{30.60}} & 0.308 & 0.181 & 38.33 \\ 

\bottomrule 
\end{tabular} 
\vspace{-5mm}
\end{table}

\subsubsection{Effectiveness of High Classifier-Free Guidance Scale}

\begin{figure*}[t!]
    \centering
    \includegraphics[width=0.8\linewidth]{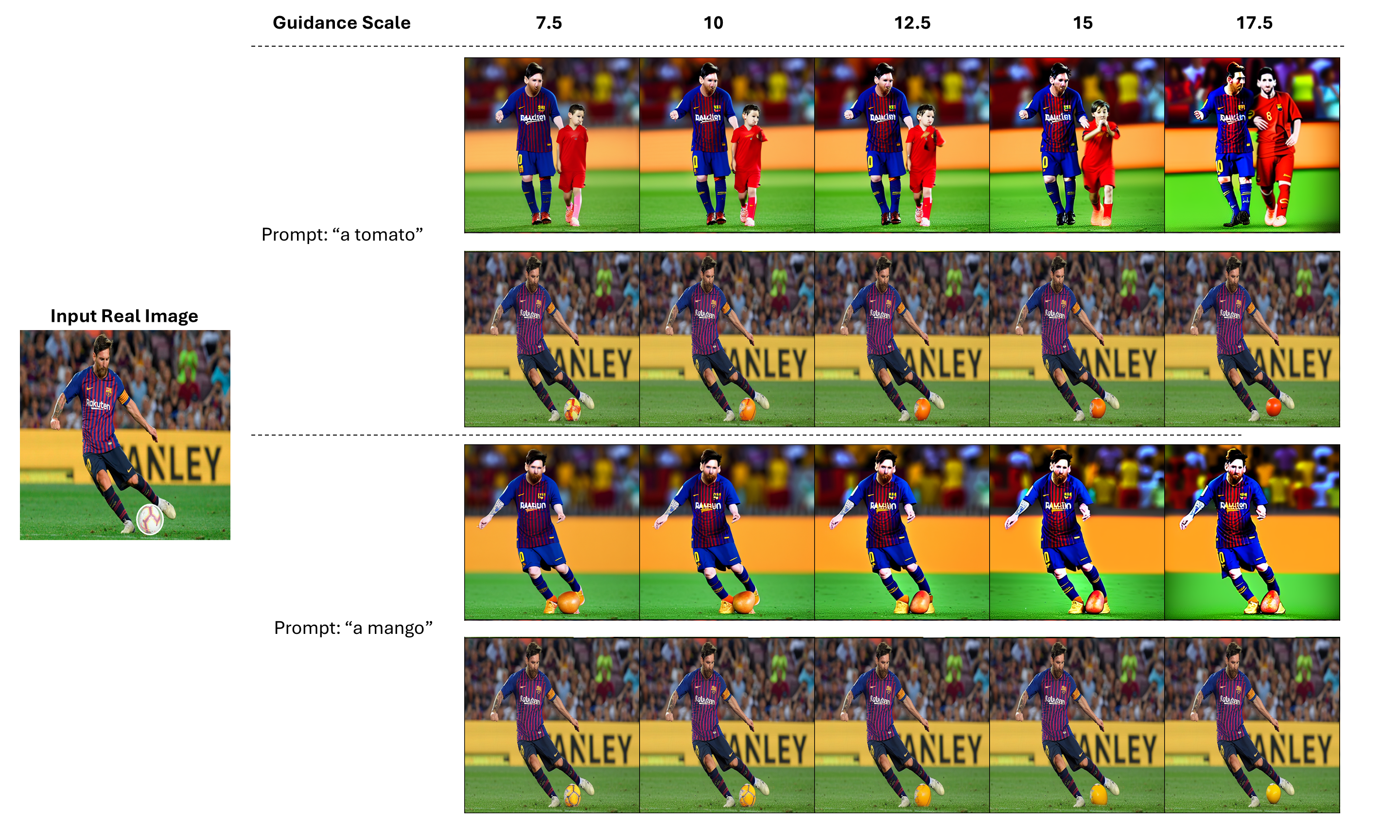}
    \caption{We demonstrate the effect of image consistency and control with and without using our method. Our approach allows for a high guidance scale, achieving the desired results without distorting the image, requiring only a simple prompt from the user.}
    \label{fig:alstudy}
\end{figure*}

Most diffusion models rely on classifier-free guidance~\cite{ho2021classifierfree}, where a low guidance scale often yields abstract or irrelevant results, while a high guidance scale may produce rigid or over-constrained images by forcing the model to adhere too strictly to the prompt, thus compromising naturalness and diversity. In contrast, CPAM applies a high guidance scale {only to the localized regions requiring modification}, rather than the entire image. This selective guidance strategy effectively reduces distortion artifacts that typically arise from over-editing globally, especially in background regions. As a result, CPAM can produce faithful edits using {simple prompts} without the need for detailed or overly specific descriptions, thereby making the editing process more intuitive (Fig.~\ref{fig:alstudy}).

\begin{figure}[t!]
    \centering
    \includegraphics[width=\linewidth]{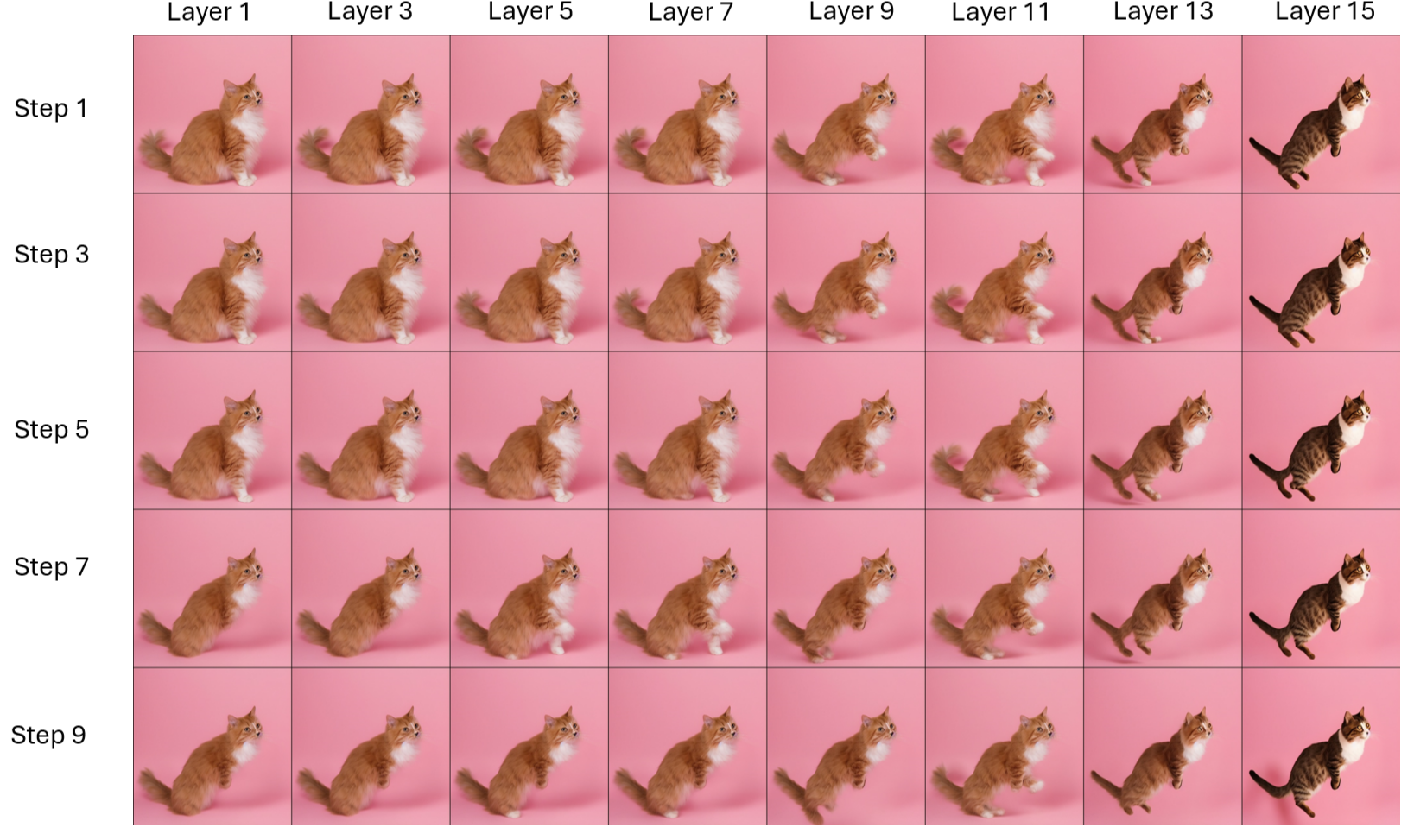}
    \caption{Step and layer scheduling. With the prompt: "A photo of a jumping cat," we control the preservation of the original cat using the preservation adaptation module. The module queries the semantic content after step $s$ and layer $l$, targeting layers 1, 3, 5, 7, 9, 11, 13, 15 in the U-Net and denoising steps 1, 3, 5, 7, 9. This configuration shows the trade-off between maintaining the original cat's features and introducing the novel pose described by the prompt.}
    \label{fig:Pose_view}
    \vspace{-3mm}
\end{figure}

\subsubsection{Step and Layer Scheduling}

\blue{We control the interaction between preservation and editing through step and layer scheduling during the diffusion process. Early denoising steps primarily establish coarse geometric structure guided by the text prompt, while later steps progressively refine fine-grained appearance details such as texture and identity. Based on this observation, querying semantic information too early tends to overly preserve features from the source image, resulting in limited changes in pose or view. In contrast, querying at very late stages allows the generated structure to dominate, often leading to outputs that deviate significantly from the original object, as illustrated in Fig.~\ref{fig:Pose_view}. To balance these effects, we select an intermediate configuration ($T=3$ and $L=7$--$9$), which provides a good trade-off between preserving semantic consistency and enabling flexible transformations. This behavior reflects a trade-off between structural preservation and editing flexibility in diffusion-based image editing. Notably, this scheduling strategy remains stable across different diffusion backbones. For SDXL, which contains a larger number of attention layers, we adopt a proportional configuration by selecting middle layers (approximately layers 30–35), maintaining the same relative position within the network.}

\subsection{User Study}
\label{us}

To assess the effectiveness of our proposed CPAM, we conducted a user study comparing it against several leading prompt-based editing methods utilizing diffusion models. The methods evaluated include SDEdit~\cite{meng2022sdedit}, MasaCtrl~\cite{cao_2023_masactrl}, PnP~\cite{Tumanyan_2023_CVPR}, FPE~\cite{liu2024towards}, DiffEdit~\cite{couairon2022diffedit}, Pix2Pix-Zero~\cite{parmar2023zeroshot}, LEDIT++~\cite{brack2024ledits++}, and the fine-tuning method Imagic~\cite{kawar2023imagic}.

\subsubsection{Metrics}

To ensure a comprehensive evaluation, we defined four key metrics: object retention, background retention, realism, and overall satisfaction. These metrics are designed to assess the methods' effectiveness in executing realistic edits while preserving important features of the original images:

\begin{itemize}
    \item \textit{Object Retention}: This metric evaluates how well the method preserves the identity and details of the main object in the image during editing.
    
    \item \textit{Background Retention}: This assesses the method's ability to maintain the integrity and appearance of the background while altering the main object.
    
    \item \textit{Realism}: This metric analyzes the realism of the edits, particularly in the context of non-rigid transformations.
    
    \item \textit{Satisfaction}: This measures the degree to which the edited image aligns with the textual description provided as the editing prompt.
\end{itemize}

\subsubsection{Participants}

\begin{figure}[t!]
    \centering
    \includegraphics[width=\linewidth]{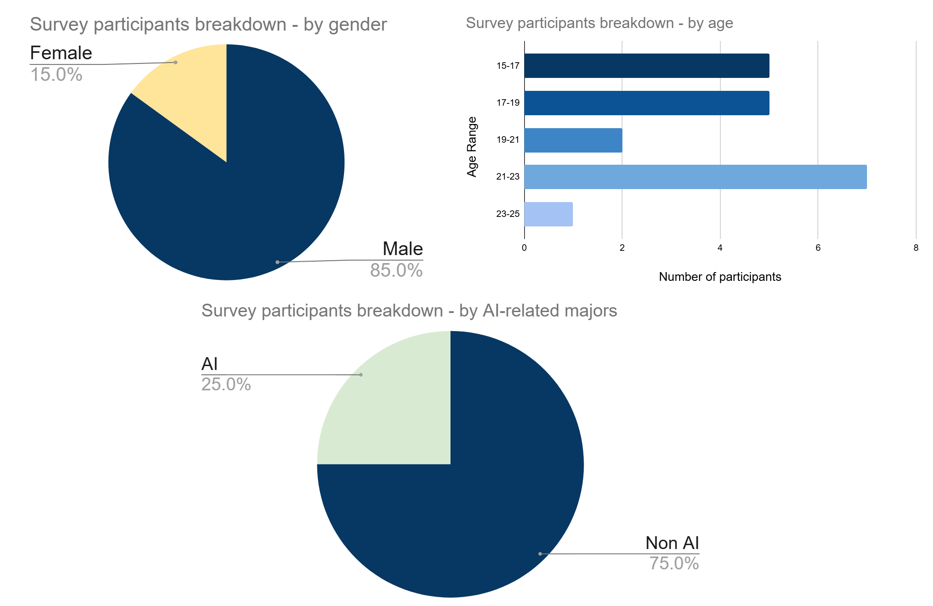}
    \caption{\blue{Information of participants.}}
    \label{fig:user_info}
\end{figure}

\blue{We invited 20 participants (17 males and three females, age from 16 to 22)} from our research community, including students with knowledge about AI and those from outside the industry, to participate in our study. All participants were new to AI generative tasks, although some had previously participated in various user studies related to AI. With diverse professional backgrounds, they brought different perspectives to the evaluation process, ensuring an objective assessment. \blue{Overview of participants' information is shown in Fig.~\ref{fig:user_info}.}

\subsubsection{Setup}

\blue{All methods were evaluated using the publicly available checkpoints of SD-1.5. We organized the participants into 20 batches, each randomly selecting 50 samples from a pool of 104 samples and shuffling the methods for evaluation. The original image and the images generated by the four methods were presented side-by-side for evaluation. To ensure objectivity, we blinded the method so that participants did not know which method the image belonged to, including our method. }

\subsubsection{Tasks}
\blue{The participants were asked to rate the performance of each of the four methods on a scale of 1 to 6 for four metrics based on their perspectives. They must follow the order of samples and methods in their batch. For samples where retaining the object was not required, participants left the rating blank in the cell corresponding to the retention of object metric.}

\subsubsection{Apparatus and Procedure}

\blue{Our user study was conducted online and in our lab, where participants completed the assigned tasks in their respective batches. The total time for these study sessions was approximately 4 hours per person. Some sessions were video-recorded for further analysis.}

\subsubsection{Quantitative Results}

\begin{table}[!t]
\centering
\caption{User study results measuring participants' opinion (1: very bad, 6: very good) in rating image editing methods. Our CPAM significantly outperforms existing methods. \textcolor{red}{\textbf{Bold}}, \textcolor{blue}{\underline{underline}} indicate the best and second best scores. Imagic~\cite{kawar2023imagic} is the fine-tuning method (FT).}
\label{tab:comparison}
\resizebox{\linewidth}{!}{
\begin{tabular}{lcccc}
\toprule
\textbf{Method}     & \begin{tabular}[c]{@{}l@{}}\textbf{Object}\\ \textbf{Retention}\end{tabular} & \begin{tabular}[c]{@{}l@{}}\textbf{Background}\\ \textbf{Retention}\end{tabular} & {\textbf{Realistic}} & {\textbf{Satisfaction}} \\
\hline

SDEdit~\cite{meng2022sdedit}     & 3.63             & 3.19                & 3.38                    & 2.42                    \\
MasaCtrl~\cite{cao_2023_masactrl}   & 4.01             & 4.17                & 4.32                    & 3.11                    \\
PnP~\cite{Tumanyan_2023_CVPR}          & \textcolor{blue}{\underline{4.61}}             & 4.49                & 4.20                    & 2.63                    \\
FPE~\cite{liu2024towards}       & 4.50             & 4.44                & 4.33                    & 2.53                    \\
DiffEdit~\cite{couairon2022diffedit}  & 4.58             & 4.57                & 4.40                    & 3.13                    \\
Pix2Pix-Zero~\cite{parmar2023zeroshot} & 2.11            & 4.23                & 1.84                    & 1.93                    \\
LEDIT++~\cite{brack2024ledits++}   & 4.38             & \textcolor{blue}{\underline{4.95}}                & \textcolor{blue}{\underline{4.57}}                    & 3.26                    \\
Imagic~\cite{kawar2023imagic} (FT)     & 3.74             & 3.48                & 4.30                    & \textcolor{red}{\textbf{4.82}}           \\
\rowcolor{lightgray} \textbf{CPAM-SD1.5 (Ours)} & \textcolor{red}{\textbf{4.72}}    & \textcolor{red}{\textbf{5.09}}       & \textcolor{red}{\textbf{4.69}}           & \textcolor{blue}{\underline{3.30}}                    \\
\bottomrule
\end{tabular}
}
\vspace{-3mm}
\end{table}

\begin{figure}[t!]
    \centering
    \includegraphics[width=\linewidth]{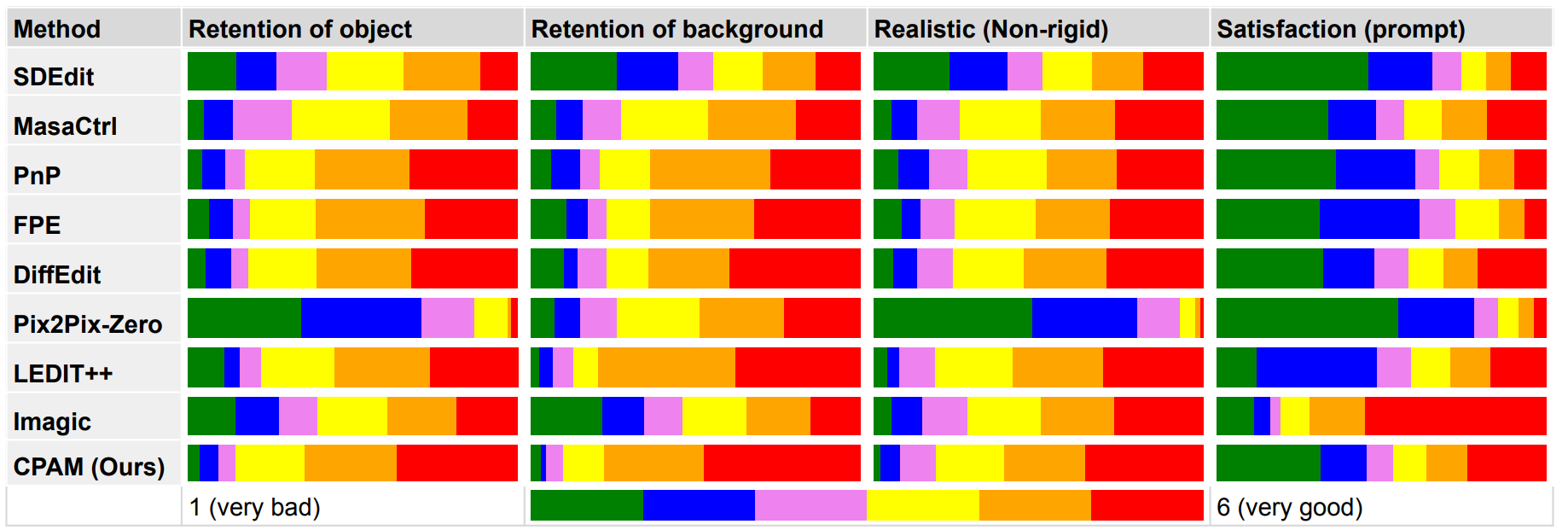}
    \caption{\blue{Statistical ratings for each method.}}
    \label{fig:statistic_ratings}
\end{figure}

Table~\ref{tab:comparison} presents the Mean Opinion Score (MOS) derived from the participants' ratings; and detailed ratings are shown in Fig.~\ref{fig:statistic_ratings}. The results demonstrate that our CPAM significantly outperforms the other methods across most metrics. Notably, CPAM received the highest ratings for object retention, background retention, and realism, indicating its superior ability to maintain key elements of the images while executing edits effectively. While Imagic~\cite{kawar2023imagic} excelled in user satisfaction but it faced challenges in background retention, occasionally produced unrealistic outputs, and required significantly more time for fine-tuning and optimization for each image-prompt pair.

Overall, the user study reinforced the findings from our qualitative and quantitative evaluations, highlighting the effectiveness of CPAM in real image editing tasks.



\subsection{Limitations and Discussion}

\begin{figure*}[!t]
    \centering
    \includegraphics[width=\textwidth]{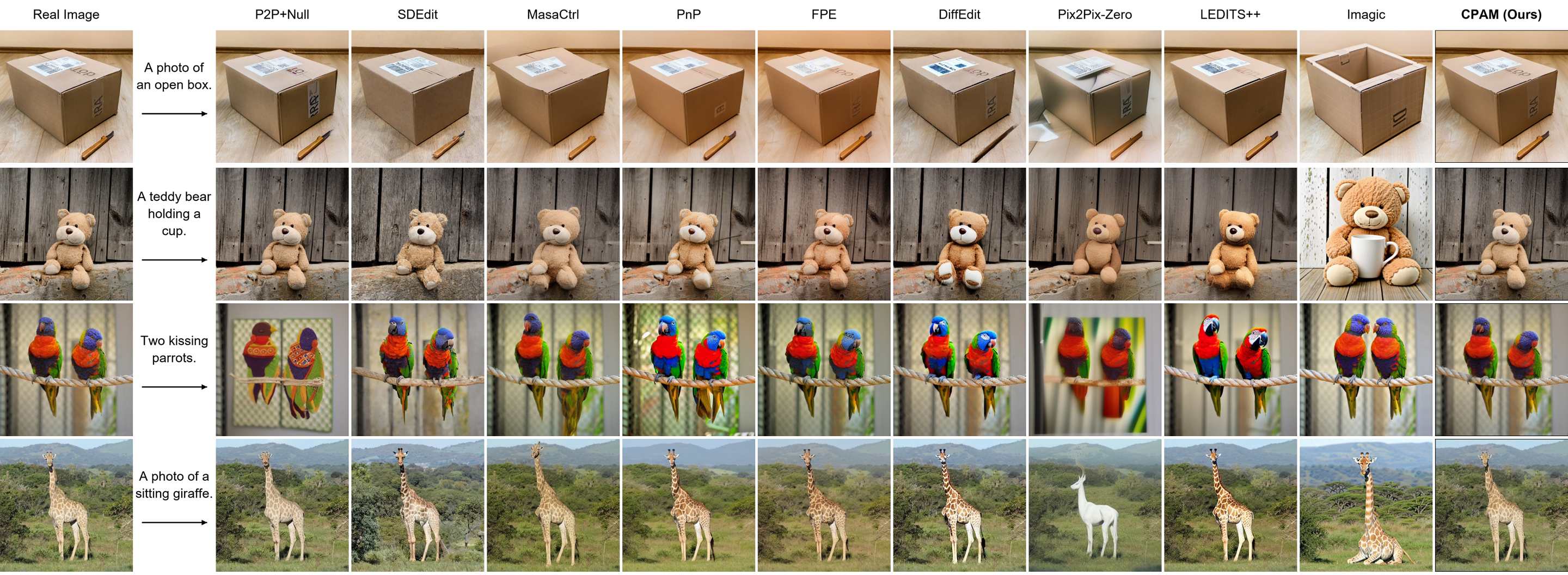}
    \vspace{-5mm}
    \caption{Without fine-tuning the model, it cannot generate novel views and poses of objects aligned with the text prompt. We compare these tuning-free methods to the fine-tuning method Imagic, which can generate novel views and poses.}
    \label{fig:failure_cases_pose_view}
    \vspace{-3mm}
\end{figure*}

\begin{figure}[t!]
    \centering
    \includegraphics[width=\linewidth]{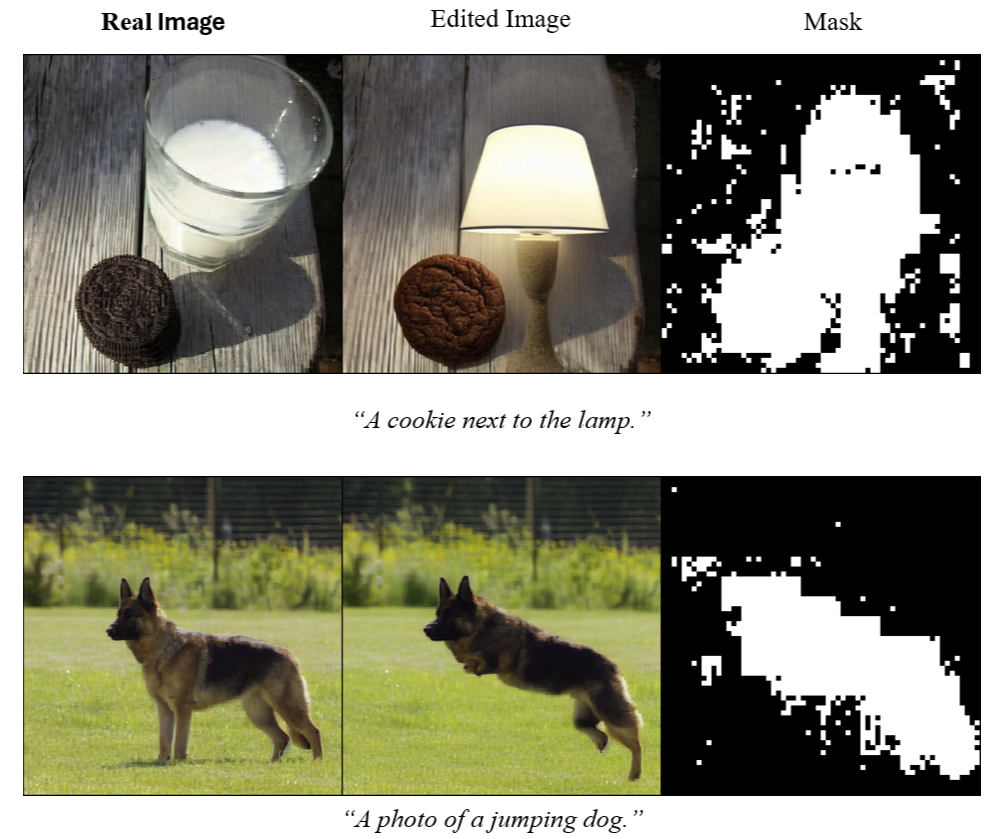}
    \caption{\blue{The failure case occurs when the mask is imprecise. In the first sample, the mask often captures all the salient objects within the image and cannot focus solely on the desired object, even if we aggregate attention maps correlated to a specific token like "lamp". In the second sample, the dog's mask loses its legs.}}
    \label{fig:failure_cases_imprecise_mask}
\end{figure}

\begin{figure*}[t!]
    \centering
    \includegraphics[width=0.83\linewidth]{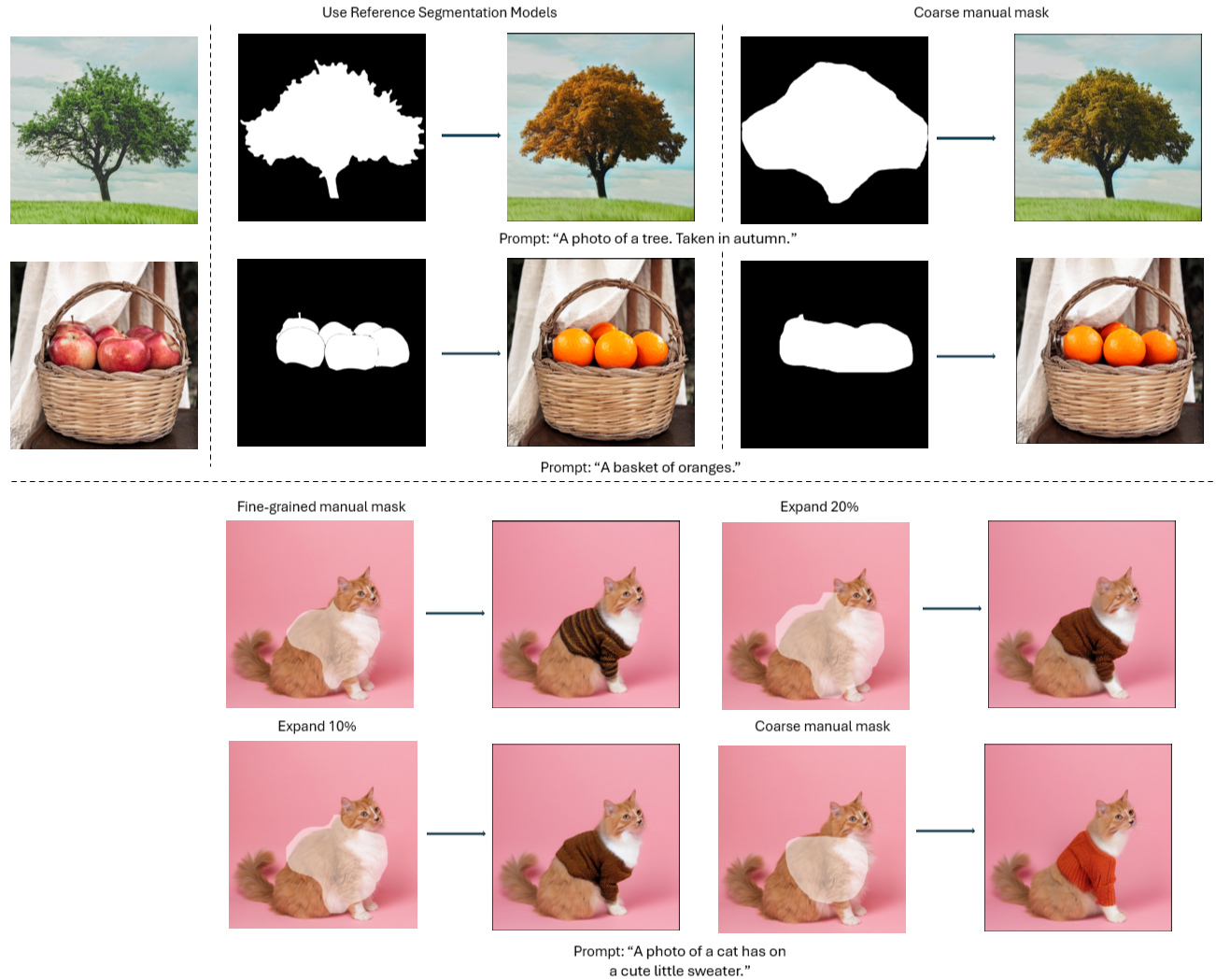}
    \caption{\blue{Illustration of various mask generation methods and their impact on the performance of our method. In the bottom subfigure, expanding the fine-grained manual mask input mask by 10 percent does not affect the output, but at 20 percent expansion, the output begins to distort, affecting the cat's left eye. With a coarse manual mask, the sweater changes to a red color, demonstrating how mask accuracy influences the results.}}
    \label{fig:accuracy_mask}
\end{figure*}

\blue{\bluex{\textbf{Pre-Trained Model Capacity Constraints.}} Like other zero-shot diffusion editing methods, CPAM inherits certain limitations from its underlying pre-trained diffusion model. Because the framework manipulates internal attention representations without requiring additional training, its overall performance is strictly bounded by the base model's capacity. Consequently, \bluex{when significant modifications in pose or viewpoint are requested, the generated outputs may occasionally fail to perfectly align with the target prompt or achieve the desired structural transformations} (see Fig.~\ref{fig:failure_cases_pose_view}).}

\bluex{\textbf{Dependence on Target Mask Estimation.} To guide spatial control during editing, CPAM relies on a source mask ($M_s$) that explicitly separates the object from its background. A target mask ($M_t$) is subsequently estimated during synthesis to track the object's evolving structure. In practice, $M_t$ is either derived directly from $M_s$ through simple geometric operations (\textit{e.g.}, mask expansion, extending the convex hull) or generated automatically by aggregating cross-attention maps across various denoising steps and layers. For minor pose or viewpoint adjustments (\textit{e.g.}, a dog facing right edited to face left), where the edited object typically remains near the original location, extending the convex hull of $M_s$ is generally sufficient to cover newly exposed areas. Conversely, for major structural transformations (\textit{e.g.}, a sitting dog edited into a jumping dog), tracking the new object boundaries requires aggregating cross-attention maps to estimate \(M_t\).}

\bluex{However, this reliance on target mask estimation can introduce failure cases, particularly when object localization becomes unreliable due to substantial structural changes, inaccurate masks, or attention drift during the tracking process. Additionally, because the attention-based mask estimation is relative, it may not always perfectly capture the object structure, particularly when the object undergoes significant geometric changes. Furthermore, in scenes containing multiple objects, the estimated attention may occasionally drift toward unintended instances, resulting in less precise object localization and editing.}

\blue{\bluex{\textbf{Failure Cases from Inaccurate Masks. }} Because of mask dependency, inaccurate source masks can directly lead to editing failures. \bluex{For example, if the mask misses certain object parts, such as the legs of the dog in Fig.~\ref{fig:failure_cases_imprecise_mask}, the generated result may lose these regions during editing. Alternatively, the mask may include multiple salient objects in the scene rather than focusing solely on the intended target, which may cause unintended edits.} These challenges underscore the critical need for reliable mask estimation in spatially controlled editing, a limitation shared by most mask-guided frameworks.}

\blue{\textbf{Robustness to Mask Imperfections.} Despite these dependencies, experiments demonstrate that CPAM remains reasonably robust to moderate mask inaccuracies. As illustrated in Fig.~\ref{fig:accuracy_mask}, the method can still produce meaningful edits when using imperfect masks, such as coarse manual masks or masks generated by off-the-shelf segmentation tools (\textit{e.g.}, SAM~\cite{kirillov2023segment}). Specifically, minor boundary deviations, such as a 10\% mask expansion, do not significantly degrade the editing quality. However, more substantial deviations can introduce artifacts or distort the object's appearance, highlighting the natural sensitivity of spatial editing techniques to mask precision}

\blue{\bluex{\textbf{Small Objects Editing Challenges.}} Editing very small objects poses a distinct challenge due to the training data distribution of large diffusion models, which typically emphasize highly salient objects over minor spatial details. As a result, \bluex{modifying content in small or less prominent regions can be unstable.} Nevertheless, our method is still able to perform edits in such regions by increasing the guidance scale to strengthen the influence of the text prompt, as illustrated in Fig.~\ref{fig:alstudy}. Moving forward, addressing these limitations holistically will likely require the integration of more accurate object localization mechanisms or the joint learning of spatial grounding alongside image editing.}

\section{Conclusion}\label{sec13}

Our CPAM facilitates various zero-shot real image editing tasks by leveraging both self-attention and cross-attention mechanisms within SD models. Overcoming existing limitations, CPAM employs a preservation adaptation process to meticulously control and retain various object attributes while preserving the background. Additionally, our method features a localized extraction module to prevent undesired effects of target prompts on non-desired spatial regions, enabling precise object editing within images. We also introduce IMBA dataset, providing rich information for comprehensive image manipulation evaluations. Empirical results demonstrate that our CPAM consistently outperforms existing leading editing techniques in achieving complicated and non-rigid edits.

\section*{Acknowledgments}

This research is funded by Vietnam National Foundation for Science and Technology Development (NAFOSTED) under Grant Number 102.05-2023.31.



\bibliographystyle{IEEEtran}
\bibliography{references}



\vspace{-30pt}
\begin{IEEEbiographynophoto}{Dinh-Khoi Vo} is currently a graduate student at the University of Science, Ho Chi Minh City, Vietnam. He received the B.Sc. degree in software engineering from the University of Science in 2024. His research interests include machine learning, computer vision, and human-computer interaction.
\end{IEEEbiographynophoto}

\vspace{-30pt}
\begin{IEEEbiographynophoto}{Thanh-Toan Do} is currently a Senior Lecturer at the Faculty
of Information Technology, Monash University. He obtained his Ph.D. in computer science at INRIA Rennes, France, in 2012. is research interests include computer vision and machine learning.
\end{IEEEbiographynophoto}

\vspace{-30pt}
\begin{IEEEbiographynophoto}{Tam V. Nguyen} is currently a professor at the University of Dayton, Ohio, United States. He received the Ph.D. degree in computer science from the National University of Singapore in 2013. His research interests include computer vision, deep learning, and mixed reality.
\end{IEEEbiographynophoto}

\vspace{-30pt}
\begin{IEEEbiographynophoto}{Minh-Triet Tran} is currently a professor at the University of Science, Ho Chi Minh City, Vietnam. He received the Ph.D. degree in computer science from the University of Science in 2009. His research interests include machine learning, computer vision, and multimedia.
\end{IEEEbiographynophoto}

\vspace{-30pt}
\begin{IEEEbiographynophoto}{Trung-Nghia Le} is currently a senior researcher and lecturer at the University of Science, Ho Chi Minh City, Vietnam. He received the Ph.D. degree in computer science from the National Institute of Informatics, Japan, in 2018. His research interests include machine learning, computer vision, and multimedia.
\end{IEEEbiographynophoto}

\vfill

\end{document}